\documentclass[11pt, a4paper, onecolumn, copyright, gr]{google}

\usepackage[numbers, sort&compress, square]{natbib}

\usepackage{graphicx}

\usepackage[expansion=false]{microtype}

\usepackage{array,makecell}

\usepackage{multirow}

\usepackage{tcolorbox}
\tcbuselibrary{breakable}
\newtcolorbox{promptbox}{
    colback=gray!5!white, 
    colframe=gray!50!black, 
    arc=2mm, 
    fontupper=\ttfamily\small,
    breakable
}
\newtcbox{\agentcmd}{on line, 
  boxsep=1pt, left=2pt, right=2pt, top=1pt, bottom=1pt,
  colback=gray!10!white, colframe=gray!40!white, 
  fontupper=\ttfamily\small,
  boxrule=0.5pt, arc=1mm}

\usepackage{tikz}
\usepackage{sansmath}
\usetikzlibrary{shapes.geometric, arrows.meta, positioning, fit, calc, backgrounds, shadows}

\definecolor{primary}{RGB}{26, 115, 232}   
\definecolor{success}{RGB}{13, 148, 136}   
\definecolor{purple}{RGB}{123, 31, 162}    
\definecolor{warning}{RGB}{245, 124, 0}    
\definecolor{darkgray}{RGB}{60, 64, 67}    
\definecolor{lightgray}{RGB}{241, 243, 244}

\tikzset{
    font=\sffamily\small,
    stagebox/.style={draw=#1, fill=#1!4, thick, rounded corners=10pt, inner sep=10pt, rectangle},
    toolbox/.style={draw=#1!80, fill=white, thick, rounded corners=6pt, inner sep=8pt,
        minimum width=4.8cm, minimum height=0.85cm, rectangle, align=center,
        font=\sffamily\footnotesize,
        drop shadow={opacity=0.04, shadow xshift=1pt, shadow yshift=-1pt}},
    externalbox/.style={draw=#1, fill=#1!8, thick, rounded corners=8pt, inner sep=10pt,
        minimum width=5.0cm, rectangle, align=center, font=\sffamily\footnotesize,
        drop shadow={opacity=0.06, shadow xshift=1pt, shadow yshift=-1pt}},
    arrow/.style={->, >={Stealth[scale=1.2]}, thick, draw=darkgray},
    widearrow/.style={->, >={Stealth[scale=1.4]}, very thick, draw=#1},
    dashedarrow/.style={->, >={Stealth[scale=1.2]}, dashed, thick, draw=#1},
    stagetitle/.style={font=\sffamily\large\bfseries, text=#1, align=center},
    stagesubtitle/.style={font=\sffamily\scriptsize\bfseries, text=darkgray, align=center},
    handlelabel/.style={fill=white, draw=darkgray!40, thick, rounded corners=4pt, inner sep=6pt,
        font=\sffamily\scriptsize, align=center,
        drop shadow={opacity=0.04, shadow xshift=1pt, shadow yshift=-1pt}}
}

\uselogo{}

\title{Planetary Prediction Engine: Autonomous Geospatial Prediction via Intelligent Data Selection and Foundation Model Embeddings}

\author[*,1]{Evelyn Ma}
\author[*,\textdagger,1]{Rama Kumar Pasumarthi}
\author[1]{Kishwar Shafin}
\author[1]{Mandar Sharma}
\author[1]{Mimi Sun}
\author[1]{Hamed Sadeghi}
\author[2]{Dav M. Ebengo}
\author[2]{Onesime Mbulayi}
\author[3]{Ciara Judge}
\author[1]{Rouslan Solomakhin}
\author[1]{John Wamburu}
\author[1]{William Ogallo}
\author[1]{Aisha Walcott-Bryant}
\author[1]{Sanxing Chen}
\author[1]{Arbaaz Muslim}
\author[1]{Yael Mayer}
\author[1]{Ronald Ho}
\author[1]{Roy Lee}
\author[1]{Ruth Alcantara}
\author[1]{Abdoulaye Diack}
\author[1]{Monica Bharel}
\author[1]{Lambert Rosique}
\author[1]{Jeremy Amez-Droz}
\author[1]{Christopher Haire}
\author[1]{James Manyika}
\author[1]{Yossi Matias}
\author[1]{Niv Efron}
\author[**,1]{Gautam Prasad}
\author[**,1]{Shravya Shetty}

\affil[*]{Equal first authors (alphabetical order)}
\affil[**]{Co-last authors}
\affil[1]{Google Research}
\affil[2]{Institut National de Recherche Biomédicale, Democratic Republic of Congo}
\affil[3]{University of Oxford}

\correspondingauthor{ramakumar@google.com, sshetty@google.com}

\begin{abstract}
{\small 
Addressing critical global challenges, from food security and disaster risk to disease outbreaks and socio-economic vulnerability, demands high-fidelity geospatial modeling.
However, building predictive planetary models remains bottlenecked by a fragmented data ecosystem, requiring manual data retrieval, multimodal data curation and fusion along with iterative model selection.
As part of Google Earth AI, we present the \textbf{Planetary Prediction Engine (PPE)}, an autonomous AI system that executes this end-to-end workflow directly from natural-language queries.
PPE synthesizes multimodal datasets on the fly, retrieving spatiotemporally relevant covariates across open-web and Earth observation platforms (Data Commons, Google Earth Engine) and fusing them with geospatial foundation model embeddings (PDFM, AlphaEarth).
Simultaneously, it searches over task-tailored model architecture families with automated overfitting guards.
Across diverse tasks, geographies, and scientific domains, PPE consistently outperforms state-of-the-art or manually tuned expert baselines.
For US spatial regression, PPE improves mean $R^2$ across 21 CDC health indicators (76.8\% vs.\ 60.0\%), FEMA national risk indices (64.9\% vs.\ 60.0\%), and the Social Vulnerability Index (66.2\% vs.\ 58.6\%).
For spatial downscaling in data-scarce settings, PPE integrates localized proxies to double baseline accuracy in Nigerian food security indicators ($R^2$ of 66.1\% vs.\ 31.5\%).
For epidemiological nowcasting of the 2026 DRC Bundibugyo Ebola outbreak, PPE achieves a Recall@10 of 83.3\% (identifying 15 of 18 newly invaded health zones across five weekly forecasts), a {+10.3} percentage-point improvement over the public state-of-the art modeling ($\sim$73\%).
By combining autonomous multimodal planetary data discovery with targeted model optimization, PPE lowers the technical barrier to planetary-scale analytics, enabling rapid, customized, expert-level deployment.
}
\end{abstract}

\begin{document}

\maketitle

\section{Introduction}


\begin{figure*}[t]
    \centering
     \includegraphics[width=\linewidth]{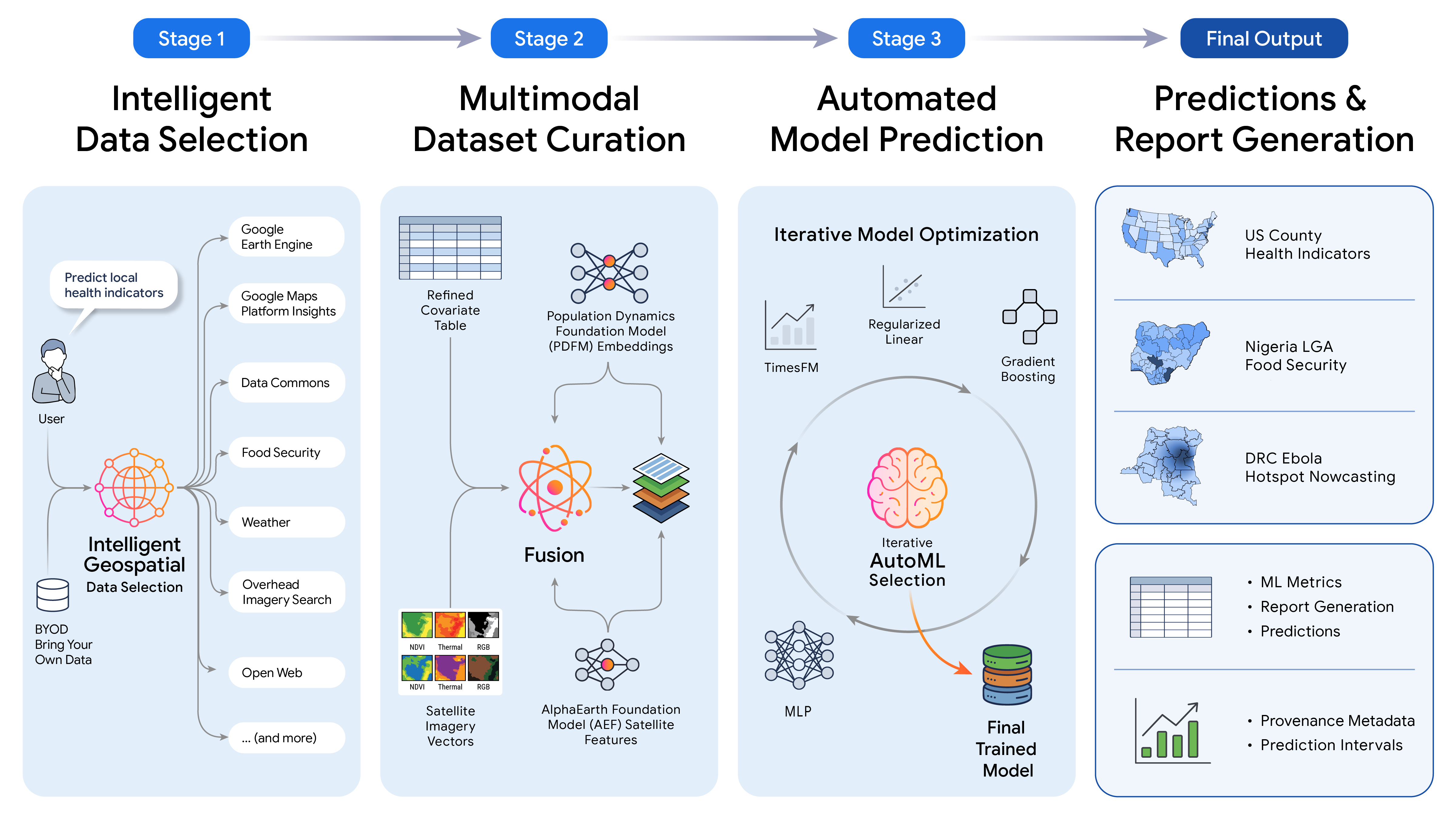}
    \caption{\small The \textbf{Planetary Prediction Engine}’s end-to-end workflow. The system decomposes the predictive workflow into three modular stages: (1) Intelligent data selection, (2) dataset curation, and (3) AutoML \& prediction to produce the  predictions and report. Off-the-shelf LLMs serve as orchestrators within each stage.  }
    \label{fig:architecture}
\end{figure*}

\subsection{Motivation and Challenges in Planetary Analytics}

Addressing pressing global crises such as mapping socioeconomic inequality, mitigating regional food insecurity, and tracking active disease outbreaks requires high-fidelity, real-time geospatial modeling. Accurate predictions allow humanitarian organizations, governments, and public health agencies to make critical, time-sensitive decisions. For instance, during an active viral outbreak, predicting spatial transmission corridors enables public health organizations to prioritize vaccine distribution and deploy mobile clinics to high-risk zones. Similarly, downscaling food consumption indicators to local administrative units allows agricultural agencies to target relief efforts to vulnerable communities \citep{funk2015climate, burke2021satellite}. Historically, translating raw Earth observations and demographic data into these downstream interventions has relied on distinct, manual modeling paradigms. These include spatial regression and super-resolution downscaling to estimate socioeconomic indicators from satellite imagery \citep{jean2016combining, yeh2020using}, as well as epidemiological nowcasting and spatial transmission modeling to correct for surveillance lags and simulate anisotropic transmission via Bayesian smoothing \citep{mcgough2020nowcasting} and spatial flux equations \citep{simini2012universal, wesolowski2012quantifying}.
However, executing these predictive workflows is notoriously labor-intensive, creating a severe bottleneck when rapid deployment is critical. For example, a geospatial predictive modeling workflow for epidemiological nowcasting can have more than 700 steps across data selection, curation and model optimization. For any given socio-environmental task, specialized teams must manually navigate a highly fragmented geospatial data ecosystem. They must conduct rigorous domain research to identify relevant proxy signals, retrieve and clean data from decentralized public and private repositories, and fuse these covariates with domain-specific geospatial foundation models \citep{agarwal2024general, clay2024foundation,jakubik2023prithvi}. These models, such as the Population Dynamics Foundation Model (PDFM)\citep{agarwal2024general} for demographics and AlphaEarth \citep{brown2025alphaearth} for satellite-derived land-use semantics, yield expressive, high-dimensional embeddings that encode latent socio-environmental patterns. Finally, researchers must design model architectures that handle spatial dependencies while enforcing strict validation guardrails to prevent spatial target leakage \citep{roberts2017cross, meyer2019importance}. Because this pipeline requires deep domain expertise and significant manual engineering, developing these models can take weeks, hindering immediate humanitarian and policy response.

Existing automated machine learning (AutoML) frameworks \citep{feurer2015efficient, wang2021flaml} are poorly suited to address these challenges.
Recent AI-driven scientific discovery systems such as Empirical Research Assistance (ERA) \cite{aygun2026era} and AlphaEvolve \cite{novikov2025alphaevolve} have achieved expert-level results across scientific domains by coupling LLMs with program search to optimize a pre-specified quality metric; however, they require a well-defined objective and curated task formulation.
Similarly, while recent work has explored LLM-based autonomous agents for scientific discovery, data science, automating code generation, iterative debugging, and hypothesis formulation \citep{wu2023autogen, guo2024dsagent, hong2024data, lu2024aiscientist}, these frameworks are typically restricted to executing code on clean, pre-curated tabular datasets or software engineering tasks. They lack the specialized capabilities needed to autonomously discover and curate large scale geospatial data on the fly, align and fuse heterogeneous geospatial embedding vectors (PDFM, AlphaEarth) via research-grounded workflows, or parameterize mechanistic epidemiological models. Consequently, planetary-scale analytics remains heavily restricted by the availability of specialized human engineering teams, leaving critical prediction tasks in data-scarce or crisis-prone regions under-addressed. We defer thorough discussion on related work to Appendix~\ref{sec:appendix_related_work}.

\subsection{The Planetary Prediction Engine Framework}
To address these limitations, we propose the Planetary Prediction Engine (PPE), an autonomous AI architecture that translates a natural-language query into an executed geospatial model (Figure~\ref{fig:architecture}). A central goal of the Planetary Prediction Engine is rapid and accessible model building: by abstracting away data discovery, feature fusion, and hyperparameter tuning, the system enables non-experts to instantly build predictive models for their specific objectives. The Planetary Prediction Engine features three modular stages: Intelligent Data Selection, Multimodal Dataset Curation, and Automated Model Building and Prediction. Guided by LLM orchestrators, the system dynamically conducts signal discovery, retrieving raw covariates from both the open web and selecting appropriate Earth AI tool calls within each stage's predefined tool set.

Importantly, task-type inference occurs early in Stage 1 during the initial query parsing: the system identifies the predictive paradigm (e.g., spatial regression, super-resolution, spatial transmission, or epidemiological nowcasting) from the user's natural-language prompt and propagates these task-specific constraints to all downstream data retrieval sub-stages. This ensures that, for example, spatial transmission tasks trigger specialized mobility matrix construction and point-of-interest distance computation, while standard spatial regression tasks focus on static demographic covariate retrieval. 

We benchmark the system across a multidimensional matrix covering varying geospatial predictive modeling tasks (spatial regression, super-resolution, epidemiological nowcasting), geographies (Global North, Global South), and thematic domains (public health, environmental vulnerability, epidemiology).

The system uses frontier large language models as black-box orchestrators within each stage. The LLMs serve as natural-language interfaces for interpreting user queries and selecting appropriate tool calls within each stage's predefined tool set. See Section~\ref{sec:methods} for more details.

\subsection{Summary of key contributions}

Our primary methodological contributions are:

\begin{itemize}
\item \textbf{Autonomous Modular Agent Architecture:} A fully automated end-to-end agent architecture for geospatial prediction and epidemiological nowcasting that, given a natural-language query, performs research-grounded signal discovery to dynamically fetch and prioritize multi-source data, performs feature engineering, and trains and evaluates models without manual intervention.  
\item \textbf{Task Auto-Identification \& Objective Alignment:} Automatic identification of geospatial predictive tasks (spatial regression, super-resolution downscaling, spatial transmission modeling, epidemiological nowcasting) from high-level human specification, allowing for AutoML optimization over the agent-defined objective. 
\item \textbf{Intelligent Data Selection \& Multimodal Curation:} An intelligent data selection protocol that combines heterogeneous geospatial embeddings (PDFM, AlphaEarth) with on-the-fly intelligently selected statistical covariates, systematically exploring feature combinations that a human practitioner might overlook, while enforcing strict Automated Target Leakage Mitigation to prevent target leakage.
\item \textbf{Automated Model Building \& Prediction:} An iterative model selection protocol that searches over diverse model families (regularized linear models, gradient-boosted trees, extreme gradient boosting, multi-layer perceptrons) and hyperparameters, incorporating a multi-layered Overfitting Guard and Self-Correction loop to select the optimal model to produce final test-set predictions.
\end{itemize}
Our empirical findings demonstrate expert-level performance across this matrix:

\begin{itemize}
\item \textbf{Epidemiological Transmission Prediction (2026 DRC Ebola Outbreak)}: For real-time prediction of new disease transmission hotspots during the \href{https://www.who.int/emergencies/disease-outbreak-news/item/2026-DON614}{May–July 2026 Bundibugyo ebolavirus outbreak}, the Planetary Prediction Engine achieves a Recall@10 of 83.3\%, correctly identifying 15 of 18 newly invaded health zones across five sequential weekly forecasts. This represents a +10.3 percentage point absolute improvement over the published state-of-the-art Bayesian modeling baseline~\citep{epidemiological2026bundibugyo} (73\%), driven by fusing epidemiological signals with PDFM embeddings and intelligently selected geospatial covariates.
\item \textbf{Super-Resolution Downscaling (Nigeria Food Security and US Socioeconomics)}: For Food Security indicators, Coarse regional reporting often obscures local vulnerability. By integrating localized market shocks and microclimate anomalies, the Planetary Prediction Engine doubled baseline accuracy when downscaling food security from the provincial (ADM1) level to the Local Government Area (LGA/ADM2) level (R² 66.1\% vs. 31.5\%). For Social Vulnerability Index (SVI), Planetary Prediction Engine presents a 26.6 percentage improvement over baseline (R² 37.6\% vs. 11.0\%) 
\item \textbf{Spatial Regression (US Socioeconomics, Health, and Environmental Risks)}: For Social Vulnerability Index (SVI), the Planetary Prediction Engine achieved a mean R² of 66.2\%, delivering a 6.8 percentage point improvement over statistical covariates baselines (58.6\%). For CDC Health variables, the system achieves a mean R² of 76.8\%, significantly outperforming expert pipelines (60.0\%). 
On FEMA Environment risk metrics, Planetary Prediction Engine outperforms expert baselines on the Socioeconomic \& Composite indicators (R² 66.9\% vs. 61.1\%) and performs on par with expert benchmarks across the broader suite of environmental targets.
\end{itemize}
Across tasks, the PPE achieves 12--94\% relative $R^2$ improvement over standard baselines, from a 12\% gain on SVI spatial regression (66.2\% vs. 58.57\%) to near-doubling on Nigeria food security downscaling (66.1\% vs. 31.5\%), and a 10.3 percentage point gain on epidemiological nowcasting (Recall@10).

\section{Results}

\subsection{General Experimental Design and Problem Matrix}

To rigorously validate the adaptability and robustness of the proposed framework, we evaluate it across three distinct predictive paradigms: (1) Mechanistic Nowcasting, (2) Super-Resolution Downscaling, and (3) Spatial Regression (Table~\ref{tab:evaluation-benchmarks}). These paradigms cover a diverse spectrum of socioeconomic, public health, ecological, and humanitarian variables tailored to reflect real-world geospatial applications. Table~\ref{tab:evaluation-benchmarks} outlines the multidimensional matrix used to benchmark the system across varying machine learning tasks, geographic contexts, and thematic domains. The technical specifications, spatial granularities, and evaluation partitions for these benchmarks manifest highly differentiated distributions, reflecting dynamic real-world constraints.

\begin{table}[h]
\centering
\caption{\small Summary of Evaluation Benchmarks and Dataset Specifications. Overview of target variables, geographic scopes, and spatial resolutions across the three evaluation paradigms.}
\label{tab:evaluation-benchmarks}
\noindent\resizebox{\linewidth}{!}{
\begin{tabular}{lllllll}
\toprule[1.2pt]
\textbf{ML Paradigm} & 
\makecell[l]{\textbf{Geographic}\\ \textbf{Scope}} & 
\makecell[l]{\textbf{Thematic}\\ \textbf{Domain}} & 
\makecell[l]{\textbf{Benchmark}\\ \textbf{Dataset}} &
\makecell[l]{\textbf{Prediction}\\ \textbf{Targets}} & 
\makecell[l]{\textbf{Train / Test}\\ \textbf{Granularity}} & 
\textbf{Train / Test Size} \\
\toprule[1pt]
Nowcasting & 
\makecell[l]{Global South\\ (DRC)} & 
\makecell[l]{Humanitarian\\ Epidemics} & 
\makecell[l]{DRC Ebola \\ Outbreak Tracking} & 
\makecell[l]{one-week confirmed\\ caseloads increase} & 
\makecell[l]{Admin 3\\ health zones} & 
\makecell[l]{7 weeks $\times$ 519 zones /\\ 5 weeks $\times$ 519 zones} \\
\midrule[0.8pt]
\multirow{2}{*}{\parbox{3.2cm}{Super-Resolution Downscaling}} & 
\makecell[l]{Global South\\ (Nigeria)} & 
\makecell[l]{Food\\ Security} & 
\makecell[l]{Nigeria FCG\\ Insecurity} & FCG score & 
\makecell[l]{Admin 1 $\to$\\ Admin 2 (LGA)} & 
\makecell[l]{30 states $\times$ 40 months /\\ 581 LGAs $\times$ 40 months} \\
 & 
 \makecell[l]{Global North\\ (US)} & 
 \makecell[l]{Social\\ Vulnerability} & SVI & 5 index scores & 
 \makecell[l]{County $\to$\\ ZCTA} & $\sim$3k / $\sim$33k \\
\midrule[0.8pt]
\multirow{3}{*}{\parbox{3.2cm}{Spatial Regression}} & 
\makecell[l]{Global North\\ (US)} & 
\makecell[l]{Public\\ Health} & CDC Health & 
\makecell[l]{21 health variables} & Census Tract & $\sim$67k / $\sim$17k \\
 & 
 \makecell[l]{Global North\\ (US)} & \makecell[l]{Environmental\\ Risk} & FEMA NRI & 21 risk scores & Census Tract & $\sim$67k / $\sim$17k \\
 & 
 \makecell[l]{Global North\\ (US)} & 
 \makecell[l]{Social \\Vulnerability} & SVI & 5 index scores & County & $\sim$2.4k / $\sim$0.6k \\
\bottomrule[1.2pt]
\end{tabular}
}
\end{table}

The \textit{Nowcasting} paradigm expanded beyond stationary datasets: this module simulates the spatiotemporal trajectory of the May–July 2026 DRC Ebola Outbreak (BDBV virus) across all 519 health zones in DRC. Driven by dynamic sparsity and humanitarian urgency, the framework fuses official registry logs from National Institute of Biomedical Research (INRB)~\citep{inrb2026ebola}, OpenStreetMap infrastructure algorithms~\citep{haklay2008openstreetmap, luxen2011real}, and Earth Engine climatology forcings~\citep{gorelick2017google} (e.g., ERA5~\citep{hersbach2020era5}, WorldPop~\citep{tatem2017worldpop}). This enables stress-testing the agentic pipeline's capacity to maintain predictive fidelity during time-sensitive, anisotropic crisis events. We benchmark our model against epidemiological predictions published by INRB~\citep{mbulayi2026realtime}, and defer detailed benchmark specifications to Appendix~\ref{sec:appendix_benchmarks} and full ablation results to Appendix~\ref{sec:appendix_results_spatial_transmission}.

\textit{Super-Resolution Downscaling} explicitly engineered to validate model efficacy under multiscale administrative pooling where models must capture coarse signals to predict fine-grained targets. On the US SVI benchmark, downstream models are trained on coarse county-level covariates ($\sim 3k$ samples) but tested against high-resolution census tracts ($N\approx84k$). Broadening applicability to data-scarce Global South contexts, we predict Nigerian Food Security by estimating Food Consumption Group (FCG) metrics via dynamic pooling of World Bank price indices, rainfall/NDVI vegetation proxies, and news sentiment datasets. For the Nigeria benchmark, the system is trained at the state level (ADM1, N=30) and evaluated at the Local Government Area level (ADM2/LGA, N=581). 

\textit{Spatial Regression} applied to the Global North (United States) to address environmental hazards and chronic disease. This includes the CDC health dataset (21 health indicators, e.g., obesity, diabetes) \citep{greenlund2022places} and the FEMA National Risk Index (21 environmental indices)  \citep{fema2023nri}, both aggregating approximately 84k fine-grained census tracts under a conventional 80:20 random train/test split, which aligns with settings of public SOTA \citep{bell2025earthai}. Conversely, the Social Vulnerability Index (SVI) module predicts 5 vulnerability scores at a coarser, county-level granularity ($N\approx3k$) \citep{flanagan2011social, cdc2020svi}. 

Detailed descriptions of datasets and benchmark formulations are provided in Appendix~\ref{sec:appendix_benchmarks} and Appendix~\ref{sec:appendix_covariates_inventory}, with full benchmark evaluations provided in Appendix~\ref{sec:appendix_results}.

\textit{Baselines and Ablation Structure}. To rigorously quantify the contribution of each component, we structure our empirical evaluations across four standardized ablation tiers:

\begin{itemize}
\item Baseline / SOTA: Traditional manual expert pipelines (hand-curated features and grid-searched models by domain experts) or standalone foundation embedding baselines. 
\item PPE (Covariates): Models trained exclusively on raw statistical covariates (e.g., Data Commons, World Bank, Census) without foundation model embeddings.
\item PPE (Embeddings): Models trained on tabular covariates fused with geospatial foundation model embeddings (PDFM and/or AlphaEarth Foundation Model (AEF)). See appendix B for description on foundation model embeddings.
\item PPE (Full Stack): The complete autonomous system, incorporating Covariates + Embeddings + Intelligent Data Selection (dynamic open-web discovery, learned covariates, OSRM physical mobility networks, and nowcasting corrections where applicable).
\end{itemize}

For the predictive modeling approaches, we characterize the data preprocessing (DP) as follows: 
\begin{itemize}
    \item[(i)] Baseline DP, which executes traditional spatial aggregation, epidemiological signal processing, human mobility matrix construction, and static vulnerability scoring (see Appendix~\ref{sec:appendix_benchmarks}); to 
    \item[(ii)] Planetary Prediction Engine Automatic Tabular DP, utilizing automated, end-to-end tabular data engineering (see Section \ref{subsubsec:method-data-discrovery} and \ref{subsec:method-feature-engineering}); and ultimately to 
    \item[(iii)] Planetary Prediction Engine Automatic Full DP, which synthesizes both automated tabular engineering and deep representation alignment for multi-modal embeddings (see Section \ref{subsubsec:method-embedding-fusion} and \ref{subsec:method-feature-engineering}).
\end{itemize}

\subsection{Epidemiological Transmission and Nowcasting}

\textbf{Outbreak Hotspot Detection}. We evaluate the performance of the Planetary Prediction Engine in predicting future transmission hotspots in an outbreak, which we define as previously uninfected health zones that subsequently experience a viral spillover event. Specifically, we report Recall@10, representing the proportion of newly infected zones successfully captured within our top 10 highest-risk predictions. On a one-week rolling forecast tracking the spatial invasion of Bundibugyo ebolavirus across the Democratic Republic of the Congo (DRC), Planetary Prediction Engine achieves a Recall@10 of 83.3\%. We defer the detailed configurations of our spatial and temporal splits to Appendix ~\ref{sec:appendix_results_spatial_transmission}.

\begin{figure}
    \centering

    \includegraphics[width=0.45\linewidth]{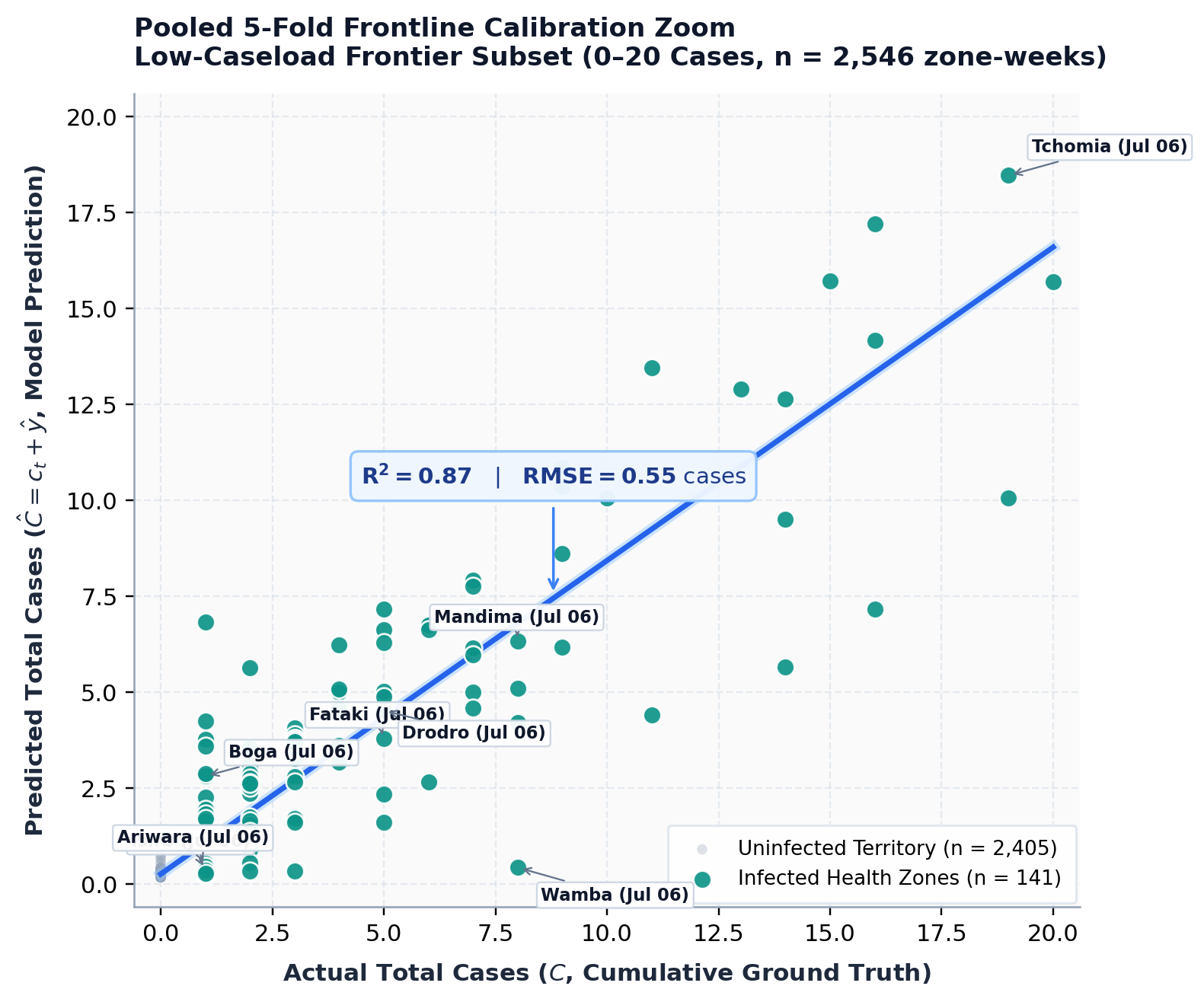}
    \includegraphics[width=0.45\linewidth]{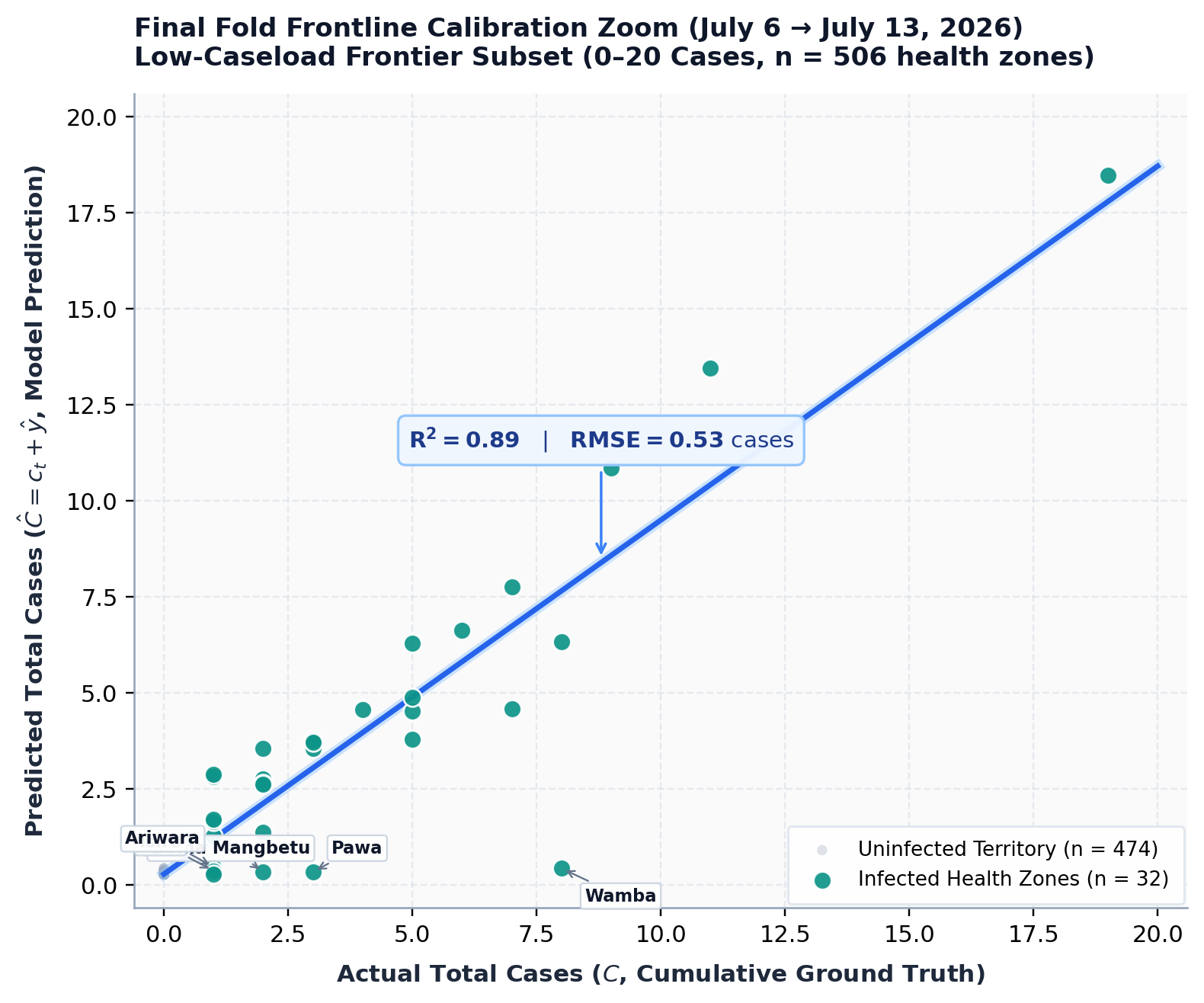}
        \caption{\small Frontline Ebola Transmission prediction visualization: Comparison of predicted total caseloads against ground-truth caseloads.}
    \label{fig:ebola_caseload}
\end{figure}

Table~\ref{tab:ebola-hotspot-recall} summarizes the hotspot detection performance (Recall@10) across feature configurations on tracking the spatial invasion of Bundibugyo ebolavirus across the Democratic Republic of the Congo (DRC). Integrating geospatial covariates with baseline epidemiological covariates improves upon the SOTA baseline (77.8\% vs. 73\%), demonstrating the utility of auxiliary geographic indicators. Peak predictive performance is achieved by the complete Planetary Prediction Engine, where fusing epidemiological signals and geospatial covariates with pre-trained PDFM embeddings yields a Recall@10 of 83.3\%. This represents a marked absolute gain of 10.3 percentage points over the state-of-the-art baseline. PPE predictions on total caseloads at frontline regions show high correlation with the ground-truth caseloads (Figure \ref{fig:ebola_caseload}). These results validate the representational power of foundation model embeddings in capturing latent demographic structure and spatial connectivity during localized spillover events. 

\begin{table}[h]
\centering
\caption{Model Performance (Recall@10) across Feature Configurations on tracking the spatial invasion of Bundibugyo ebolavirus across the Democratic Republic of the Congo (DRC).}
\label{tab:ebola-hotspot-recall}
\noindent\resizebox{\linewidth}{!}{
\begin{tabular}{lllc}
\toprule
\textbf{System} & \textbf{Feature Configuration}  &\textbf{Data Preprocessing (DP)} & \textbf{Recall@10 (\%)[95\% CI]} \\
\midrule
\makecell[l]{Baseline/ SOTA\\ (Bayesian modeling)} & Baseline Signals  & Baseline DP & $\sim 73$ \\
PPE (Covariates) & \makecell[l]{Baseline Signals +\\Geospatial Covariates} & \makecell[l]{Baseline DP +\\PPE automatic tabular DP}& 77.8 [54.8, 91.0] \\
PPE (Full Stack) & \makecell[l]{Baseline signals + \\PPE automatic full data selection} & \makecell[l]{Baseline DP + \\PPE automatic full DP} & \textbf{83.3} [60.8, 94.2] \\
\bottomrule
\end{tabular}
}
\end{table}

We defer the ablation results regarding model optimization, as well as comparison of mechanistic modeling vs spatial transmission regression to Appendix ~\ref{sec:appendix_results_spatial_transmission}.

\subsection{High-Resolution Spatial Downscaling}

In this section, we evaluate the performance of the geospatial prediction agent on Super Resolution tasks. We present performance on SVI and FCG benchmarks, covering domains of social economics and food security. We observe that the agent achieves significant outperformance over baselines.

\subsubsection{Super-Resolution Downscaling in Nigeria for Food Security Indicators}

\begin{figure}
    \centering
    \includegraphics[width=0.9\linewidth]{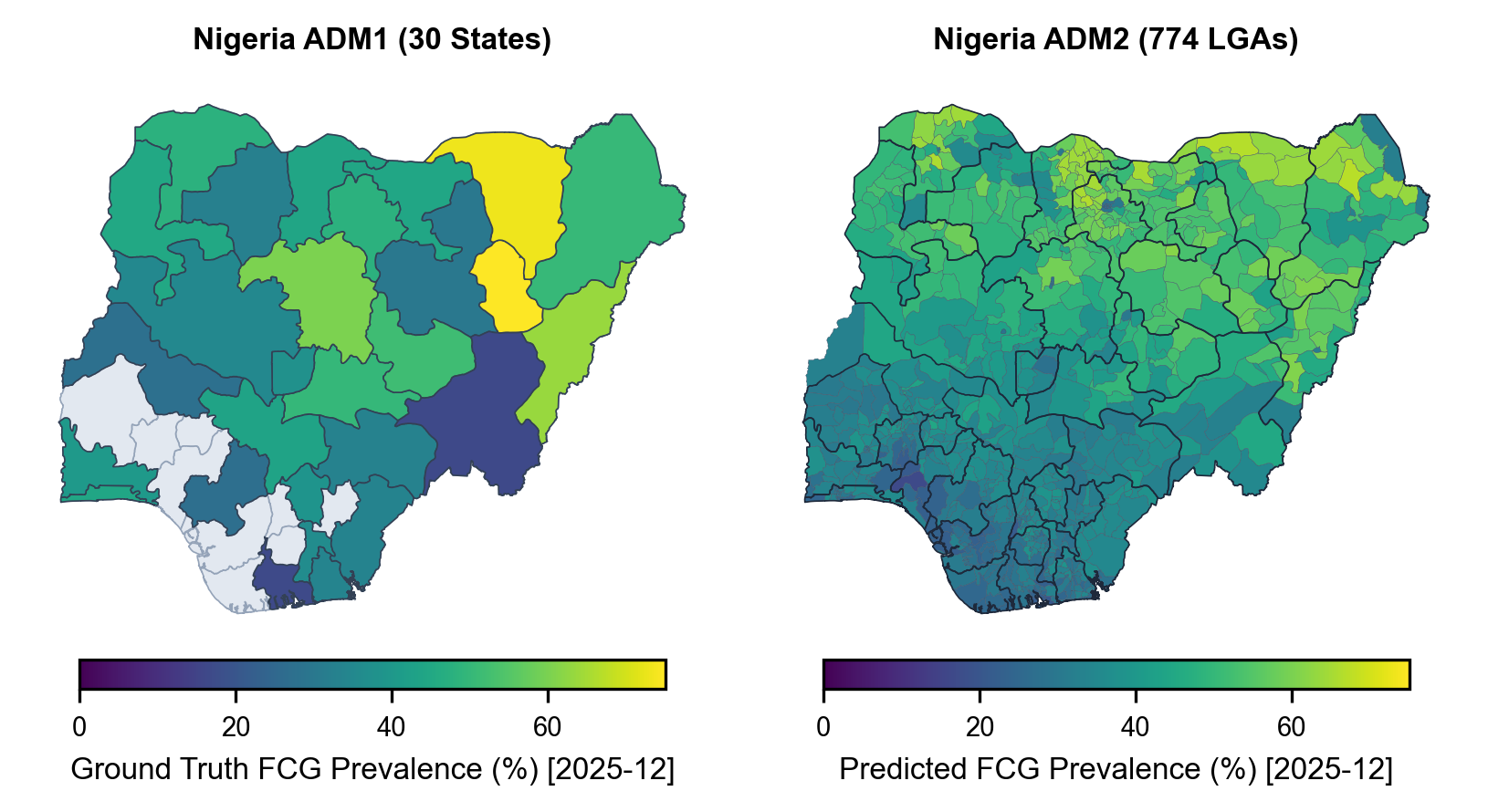}
    \caption{\small Super-Resolution Food Security Downscaling in Nigeria (ADM1 State Level $\to$ ADM2 LGA Level)}
    \label{fig:food_security}
\end{figure}

We evaluate the Planetary Prediction Engine on downscaling food security indicators across Nigeria. In developing regions, food security assessments captured by metrics such as the Food Consumption Group (FCG) \citep{wiesmann2009validation} and Integrated food security Phase Classification (IPC) \citep{ipc2021technical}. Acute food insecurity phases are typically collected through representative surveys at the coarse State / Federal Capital (ADM1) level or Local Government Area (ADM2) level. (We will describe ADM1 with State level resolution for the sake of brevity.) However, humanitarian interventions require high-resolution, fine-grained estimates at the Local Government Area (LGA) or sub-LGA level to enable better targeted interventions \citep{openshaw1984modifiable}. In our benchmark, the system is trained on ADM1 state-level data (N=30) and evaluated at the ADM2 LGA level (N=581) on a monthly basis across 40 months, representing a substantial spatial downscaling challenge.

We implement the traditional spatial-downscaling approach, Macro-Covariates + Interpolation, as the baseline. The input features consist solely of calendar month timestamps encoded as first-order Fourier seasonal harmonics $(sin(\frac{2\pi \cdot month}{12}), cos(\frac{2\pi \cdot month}{12}))$ and a linear year trend $(Year-2022)$. This formulation models the national agricultural harvest and lean cycle alongside multi-year macroeconomic drift. A gradient boosting model is trained on samples at a coarse administrative level. This baseline captures purely temporal and seasonal variance.

The Planetary Prediction Engine autonomously curates a rich multimodal feature space to drive the super-resolution downscaling model. It integrates several features including World Bank food price indices~\citep{wfp2024foodprices} (capturing localized food price index and food price inflation anomalies), World Food Programme (WFP) food insecurity metrics~\citep{wfp2021hungermap,kalkuhl2016food}, precipitation indicators \citep{funk2015climate}, Normalized Difference Vegetation Index (NDVI) indicators \citep{didan2015modis}, and VIIRS (Visible Infrared Imaging Radiometer Suite) nighttime lights (NTL) radiance~\citep{elvidge2017viirs}. These tabular data streams are fused with 330-dim PDFM~\citep{agarwal2024general} socioeconomic embeddings and 64-dim AlphaEarth satellite features~\citep{brown2025alphaearth}. 
Ground truth prevalence of insufficient food consumption was calculated as the proportion of households classified in the poor or borderline Food Consumption Group (FCG) from surveys conducted across Nigeria between August 2022 and December 2025 by the WFP, which are statistically representative at the ADM1 level. 
The household data provided by WFP is anonymized and does not contain any personal identifiable information. The coarse ADM1 FCG prevalence was used for model training and leave-one-state-out cross-validation, while localized ADM2 (LGA) ground truth was independently estimated using Multilevel Regression and Poststratification (MRP)~\citep{park2004mrp} and kept blind to the model during training. The ADM2 MRP estimates were used for out-of-fold validation, where the Planetary Prediction Engine achieved a Mean Absolute Error (MAE) of 10.0\% compared to 13.6\% for the baseline model.

Table~\ref{tab:nigeria-food-security-results} demonstrates the improved downscaling accuracy achieved by the Planetary Prediction Engine, measured using $R^2$ measured against ADM1 level ground truth data with spatial cluster bootstrap 95\% confidence interval (unit of resampling = state, k=30). While the baseline (relying on macro-covariates and basic interpolation) achieves an $R^2$ of 31.5\%, and baseline + vegetation reach 60.1\%, the complete Planetary Prediction Engine achieves an $R^2$ of 66.1\%, doubling the baseline accuracy. The Planetary Prediction Engine's intelligent data selection successfully captures agricultural seasonality, localized farming and drought shocks (via NDVI, vegetation index monthly, historical average, vegetation anomaly ratio), and urbanization and commercial infrastructure (via NTL), providing humanitarian organizations with actionable, high-fidelity vulnerability maps.

\begin{table}[h]
\centering
\caption{Model Performance ($R^2$) for Super-Resolution Food Security Downscaling (FCG) in Nigeria (ADM1 State Level $\to$ ADM2 LGA Level).}
\label{tab:nigeria-food-security-results}
\noindent\resizebox{\linewidth}{!}{
\begin{tabular}{llc}
\toprule
\textbf{System} & \textbf{Feature Configuration} & \textbf{Overall $R^2$ [95\%CI] (\%)} \\
\midrule
{Baseline} & Macro-Covariates + Interpolation & 31.5 [19.0, 39.3] \\
\multirow{2}{*}{\parbox{3.2cm}{PPE (Covariates)}} & Macro-Covariates + nighttime lights (NTL) & 49.8 [39.8, 57.9] \\
 & Macro-Covariates + vegetation & 60.1 [40.3, 72.8] \\
PPE (Full Stack) & \makecell[l]{Macro-Covariates + vegetation + NTL +\\Automatic Intelligent Selection }& \textbf{66.1} [55.9, 72.8] \\
\bottomrule
\end{tabular}
}
\end{table}

The Planetary Prediction Engine selected the following datasets:
\begin{enumerate}[label=(\roman*)]
    \item \textbf{Macro-Covariates:} Temporal features of sine and cosine cyclical month encoding $[\sin(2\pi \cdot \text{month} / 12), \cos(2\pi \cdot \text{month} / 12)]$ and temporal trend $[\text{year} - 2022]$.
    \item \textbf{Nighttime lights (NTL):} The average nocturnal visible light radiance measured by the Day/Night Band (DNB) of the VIIRS instrument aboard the Suomi-NPP satellite.
    \item \textbf{Vegetation:}
    \begin{enumerate}[label=(\alph*)]
        \item NDVI (Normalized Difference Vegetation Index): Derived from Sentinel-2 / Landsat surface reflectance using near-infrared (NIR) and red wavelengths.
        \item Vegetation Index Monthly: The monthly observed vegetation index derived from the MODIS (Moderate Resolution Imaging Spectroradiometer) sensor aboard Terra and Aqua satellites.
        \item Historical Climatological Baseline Vegetation Index: The long-term multi-year historical mean of the MODIS vegetation index for that specific calendar month.
        \item Vegetation Index Quotient / Anomaly Ratio: The ratio of (b) / (c).
    \end{enumerate}
\end{enumerate}

Planetary Prediction Engine performed a search across multiple machine learning architectures (Ridge Regression, Lasso Regression, ElasticNet, Random Forest, Gradient Boosting, XGBoost) and candidate datasets. It selected gradient boosting paired with temporal macro-covariates, NTL, and vegetation after discovering that this combination achieved the peak $R^2$ of 66.1\%, whereas incorporating additional signals resulted in lower predictive accuracy.

In the side-by-side comparison (Figure~\ref{fig:food_security}), the left map displays the ground-truth food insecurity prevalence across 30 surveyed Nigerian states in December 2025, leaving unmonitored states blank. The right map presents our super-resolution model’s predictions across all 774 LGAs nationwide, which demonstrates how the framework can bridge gaps in data-sparse regions where ground surveys are absent, but satellite telemetry is ubiquitous.

\subsubsection{Social Vulnerability Index}

We employ the standard Macro-Covariates + Interpolation approach as the baseline. A regularized linear regression model is trained on county-level targets using geographic centroid coordinates (lat,lon) alongside one-hot administrative state fixed-effects.

\begin{table}[h]
\centering
\caption{Model Performance ($R^2$) across Feature Configurations on SVI (Social Vulnerability Index) Benchmark Themes with Super-resolution tasks (from county-level to zipcode-level).}
\label{tab:svi-super-resolution-results}
\begin{tabular}{lll}
\hline
System & Feature Configuration & Mean $R^2$ [95\% CI] \\
\hline
Baseline & Macro-Covariates + Interpolation & 11.0 [10.0, 12.3] \\
PPE (Covariates) & Geospatial Covariates & 25.6 [24.3, 26.9] \\
PPE (Embeddings) & PDFM & 36.9 [36.0,37.9] \\
PPE (Full Stack) & Covariates + PDFM & \textbf{37.6} [36.7, 38.6] \\
\hline
\end{tabular}
\end{table}

To evaluate the cross-scale generalization of our framework under heterogeneous administrative boundaries, we formulate a super-resolution downscaling benchmark using the CDC's Social Vulnerability Index (SVI). Standard socioeconomic vulnerability profiles are typically aggregated and published at the coarse county level ($\sim 3,000$ in the US) to maintain statistical representative validity and prevent privacy leakages. However, localized humanitarian, epidemiological, and policy interventions require high-resolution estimates at the ZIP-code level (approximated by ZIP Code Tabulation Areas, or ZCTAs). To rigorously simulate this scenario, our downstream models are trained exclusively on coarse, county-level statistical covariates but validated prospectively against fine-grained, ZIP-code-level targets.

Table~\ref{tab:svi-super-resolution-results} summarizes the super-resolution downscaling performance ($R^2$) across feature configurations on the US SVI benchmark. Resolving sub-county variation under administrative pooling presents a severe spatial super-resolution challenge due to extreme local socioeconomic heterogeneity. Standalone modalities, i.e., raw geospatial covariates (25.6\% overall $R^2$) struggle to capture sub-county variance. A major performance breakthrough is achieved by fusing county-level covariates with our pre-trained 330-dimensional Population Dynamics Foundation Model (PDFM) embeddings, which drives the overall $R^2$ to 37.6\%. This indicates that PDFM successfully encodes latent, cross-scale socioeconomic representations that remain robust to administrative boundary pooling.

\subsection{Spatial Regression}

In this section, we evaluate the performance of the geospatial prediction agent on spatial regression tasks. We present performance on CDC health \citep{greenlund2022places}, FEMA environment risk \citep{fema2023nri}, and SVI benchmarks \citep{flanagan2011social, cdc2020svi}, covering domains of public health, environment, and social economics. Detailed benchmark definitions are deferred to Appendix~\ref{sec:appendix_benchmarks}. We observe that the agent demonstrates significant outperformance over SOTA baselines.

\subsubsection{CDC Health Variables}

Table~\ref{tab:cdc-health-results} demonstrates the lift achieved by the Planetary Prediction Engine across the ablation tiers. While the previous state-of-the-art (SOTA) manual expert pipeline achieves a mean $R^2$ of 60\%, the PPE's intelligent data selection and multimodal fusion drive mean $R^2$ to 76.8\%, a 23 percentage point improvement.

\begin{table}[h]
\centering
\scriptsize
\caption{Model Performance ($R^2$) across Feature Configurations on 21 CDC Health Variables.}
\label{tab:cdc-health-results}
\noindent\resizebox{\linewidth}{!}{
\begin{tabular}{ccc}
\toprule
\textbf{System} & \textbf{Feature Configuration} & \textbf{Mean $R^2$ [95\% CI] (\%)} \\
\midrule
Baseline / SOTA & Manual Expert Pipeline (PDFM + AEF) & 60\\
\multirow{2}{*}{\parbox{3.2cm}{PPE (Embeddings)}}  & PDFM & 59.7 [58.6, 60.9] \\
& PDFM+AEF & 61.8 [60.6, 62.9]\\
PPE (Full Stack) & PDFM + AEF + Data Commons Covariates & \textbf{76.8} [76.1, 77.6] \\
\bottomrule
\end{tabular}
}
\end{table}

\subsubsection{FEMA Environment Risk Variables}

Table~\ref{tab:fema-risk-results} evaluates the performance of Planetary Prediction Engine on 20 different 
county-level environmental and climate risk indices from FEMA against the performance 
previously reported by hand-curated models by experts \cite{bell2025earthai}.

We have split the 20 labels from FEMA into three different categories. The socioeconomic 
\& composite category contains targets that quantify human and institutional vulnerability, 
and overarching community risk. The atmospheric and climatological category contains 
hazards driven by meteorological, atmospheric and weather conditions. Finally, the 
geophysical and hydrological category covers hazards governed by geodynamic processes.

Given the FEMA target labels contain different modalities for risk assessment, from socioeconomic 
to hydrological hazard, we measured the impact of autonomous representation selection by 
conducting a systematic feature ablation study where the agent could choose the optimal 
feature configuration for each target. The agent could choose autonomously between the 
PDFM and AEF combined or in isolation, or it can combine them with Data Commons (DC) 
covariates where helpful. Overall, this autonomous feature ablation configuration 
allows Planetary Prediction Engine to achieve a higher mean $R^2$ ($61.1\%$) compared 
to the hand-curated baseline's mean $R^2$ ($59.9\%$) over all 20 labels. Specifically 
in the socioeconomic and composite category, the feature ablation configuration achieves 
$66.9\%$ mean $R^2$ compared to $61.1\%$ mean $R^2$ of the hand-curated model.

We added the full Intelligent Data Selection (IDS) pipeline to be a part of the ablation 
suite to select the winning feature suite and we observed the performance further scales 
to a nationwide mean $R^2$ of $64.9\%$ from $59.9\%$ mean from the hand-curated model.

Beyond the category aggregate, noting standout single-target gains like Social Vulnerability 
($R^2 = 67.6\%$ vs. $48.2\%$, a $+40.0\%$ gain) underscores the capability of autonomous 
multimodal feature selection and automated covariate discovery. This analysis overall 
demonstrates the generalizability of Planetary Prediction Engine to environmental and 
biophysical prediction targets.


\begin{table}[htbp]
\centering
\small
\caption{\small Model Performance ($R^2$) across Feature Configurations on FEMA Environmental Risk Variables.}
\begin{tabularx}{\textwidth}{p{4.2cm} >{\centering\arraybackslash}X >{\centering\arraybackslash}X >{\centering\arraybackslash}X >{\centering\arraybackslash}X}
\toprule
\textbf{System} & 
\begin{tabular}[c]{@{}c@{}}\textbf{1. Socioeconomic}\\ \textbf{\& Composite}\end{tabular} & 
\begin{tabular}[c]{@{}c@{}}\textbf{2. Atmospheric}\\ \textbf{\& Climatological}\end{tabular} & 
\begin{tabular}[c]{@{}c@{}}\textbf{3. Geophysical}\\ \textbf{\& Hydrological}\end{tabular} & 
\begin{tabular}[c]{@{}c@{}}\textbf{Full Nationwide}\\ \textbf{Suite}\end{tabular} \\
\midrule
Target Count & 4 & 10 & 6 & 20 \\
\midrule
\begin{tabular}[t]{@{}l@{}}PDFM + AEF\\ Manual expert\\ Mean $R^2$ (\%)\end{tabular} & 
61.1 & 64.6 & 51.1 & 59.9 \\
\midrule
\begin{tabular}[t]{@{}l@{}}PDFM + AEF + DC\\ (Feature ablation)\\ Mean $R^2$ $[95\%\text{ CI}]$ (\%)\end{tabular} & 
\begin{tabular}[c]{@{}c@{}}66.9\\ {[}66.2, 67.5{]}\end{tabular} & 
\begin{tabular}[c]{@{}c@{}}64.3\\ {[}63.6, 65.0{]}\end{tabular} & 
\begin{tabular}[c]{@{}c@{}}51.7\\ {[}49.3, 53.7{]}\end{tabular} & 
\begin{tabular}[c]{@{}c@{}}61.1\\ {[}60.2, 61.7{]}\end{tabular} \\
\midrule
\begin{tabular}[t]{@{}l@{}}PDFM + AEF + DC +\\ Intelligent Data\\ Selection\\ Mean $R^2$ $[95\%\text{ CI}]$ (\%)\end{tabular} & 
\begin{tabular}[c]{@{}c@{}}\textbf{69.4}\\ {[}68.8, 70.0{]}\end{tabular} & 
\begin{tabular}[c]{@{}c@{}}\textbf{68.3}\\ {[}67.5, 68.9{]}\end{tabular} & 
\begin{tabular}[c]{@{}c@{}}\textbf{56.2}\\ {[}54.1, 57.8{]}\end{tabular} & 
\begin{tabular}[c]{@{}c@{}}\textbf{64.9}\\ {[}64.1, 65.5{]}\end{tabular} \\
\bottomrule
\end{tabularx}
\label{tab:fema-risk-results}
\end{table}

Full results including which features were used with target-specific R² is reported in Appendix~\ref{fema-predictions}. An example prompt on how the ablation was performed is also provided in Appendix~\ref{subsubsec:appendixfemaprompt}.

\subsubsection{Social Vulnerability Index (SVI) Spatial Regression}

To maintain methodological consistency with existing expert spatial regression benchmarks on CDC health and FEMA risk indicators\citep{bell2025earthai}, we establish the SVI baseline as the optimal performance attained across all unimodal embedding-only configurations.

Table~\ref{tab:svi-regression-results} summarizes predictive R² across feature configurations, highlighting the benefits of multimodal fusion for social vulnerability modeling. Among standalone modalities, latent PDFM embeddings (R² = 58.6\%) outperform both explicit covariates (51.6\%) and AEF signals (45.1\%). Ultimately, the complete Planetary Prediction Engine pipeline (Covariates + PDFM + AEF + Intelligent Selection) achieves peak performance (R² = 66.2\%). 

\begin{table}[h]
\centering
\caption{\small Model Performance ($R^2$) across Feature Configurations on SVI Benchmark Themes (Spatial Regression at County-Level).}
\label{tab:svi-regression-results}
\begin{tabular}{llcc}
\toprule
{System} & {Feature Configuration} & {Mean $R^2$ [95\% CI] (\%)} \\
\midrule
Baseline & Foundation Model Signals & 60.3 [55.2, 64.9] \\
PPE (Covariates) & Geospatial Covariates & 50.2 [44.2, 55.6] \\
PPE (Full Stack) & Covariates + PDFM & {66.2} [61.6, 70.4] \\
\bottomrule
\end{tabular}
\end{table}

This significant margin confirms the synergistic value of the full stack of synthesizing structured socio-demographic contexts, latent spatial semantics, and physical indicators to capture the multi-dimensional complexity of geospatial vulnerability. 

\section{Discussion}
Based on analysis across spatial regression, super-resolution and epidemiological nowcasting, we look at key insights across these evaluations, along with limitations and future work.

\subsection{Key Insights}

Across the three tasks, three major insights emerge from our evaluation:

\textit{Multimodal fusion improves prediction quality.} Across all benchmarks, the combination of explicit geospatial covariates with latent foundation model embeddings (PDFM, AlphaEarth) outperforms either modality in isolation. This confirms that pre-trained geospatial representations encode complementary information to traditional statistical covariates, and that their fusion, when mediated by appropriate feature engineering and leakage prevention, yields robust predictive gains.

\textit{Autonomous data curation closes the expertise gap.} The Intelligent Data Selection pipeline, comprising grounded signal discovery, multi-repository retrieval, open-web search, and provenance-first prioritization, enables the system to assemble rich, task-specific covariate sets that rival or exceed those constructed by domain experts. This is particularly impactful in data-scarce settings (e.g., Nigeria food security, DRC Ebola), where the agent discovers localized proxy signals that would require substantial domain knowledge to identify manually.

\textit{Co-optimization of data and models is essential.} Our ablation studies consistently show that neither intelligent data curation nor automated model selection alone achieves peak performance. The multiplicative interaction between what the model sees (curated, high-fidelity features) and how it learns (optimized architectures and hyperparameters) is the primary driver of the Planetary Prediction Engine's performance advantage.

\subsection{Multimodal Synergy in Spatial Regression}

Our spatial regression evaluations in data-rich environments demonstrate a consistent performance lift across public health, environmental, and socioeconomic domains. On the 21 CDC health indicators, the Planetary Prediction Engine leverages multimodal fusion and autonomous model selection to achieve a mean R² of 76.8\%, significantly outperforming the manual expert pipeline baseline of 60\%. Similar improvements are observed for FEMA environmental risk variables, where the Planetary Prediction Engine reaches a mean R² of 64.9\% compared to the baseline of 60\%. For the Social Vulnerability Index (SVI), fusing explicit geospatial covariates with latent foundation model embeddings unlocks peak performance (mean R² of 66.2\% across 5 themes), confirming that combining structured socio-demographic indicators with dense spatial representations yields a more robust understanding of regional vulnerability than standalone modalities.

\subsection{Cross-Scale Generalization and Noise-Resolution Trade-offs}

The super-resolution downscaling benchmarks validate the system's ability to maintain predictive fidelity across heterogeneous administrative granularities. In data-scarce regions such as Nigeria, the Planetary Prediction Engine dynamically curates localized proxies, including food price anomalies and news sentiment, double the baseline accuracy when mapping food security at the LGA level (R² of 66.1\% vs. 31.5\%). However, cross-resolution projection \citep{atkinson2013downscaling} also uncovers an important noise-resolution trade-off. In the SVI super-resolution task from county to ZIP code level, adding high-resolution AlphaEarth Foundation (AEF) physical features to the covariate-PDFM stack causes a drop in overall performance (R² of 40.1\% vs. 52.0\%). This suggests that while satellite-derived terrain and land-use attributes provide rich localized context, they can introduce high-frequency noise or trigger spurious correlations at fine scales, potentially disrupting the generalization of broader socioeconomic proxies during administrative downscaling.

\subsection{Trade-offs in Epidemiological Transmission and Nowcasting}

Our experimental results demonstrate that combining intelligent data selection with automated model selection improves hotspot prediction accuracy. Relying solely on raw geospatial and dynamic epidemiological covariates yields sub-optimal performance. Integrating pre-trained PDFM embeddings with geospatial covariates provides an intermediate lift (Recall@10 of 77.8\%), but performance remains bottlenecked by the passive ingestion of all features and non-optimized model configurations. The Planetary Prediction Engine agent resolves these limitations by jointly optimizing the feature input space (via Intelligent Data Selection) and the hypothesis space (via Model Search). This co-design drives Recall@10 to 83.3\%, marking a 10.3 percentage point absolute gain over the published SOTA (~73\%).  Also, see  Appendix~\ref{sec:appendix_results_spatial_transmission} for comparison between spatial transmission modeling and epidemiological nowcasting in the context of the 2026 Ebola outbreak, which shows that spatial transmission modeling is better suited for predicting new hotspots compared to standard mechanistic SEIR models.

\subsection{Limitations and Future Work}

Firstly, the system currently relies on foundation model embeddings (PDFM, AlphaEarth) as frozen feature extractors; end-to-end fine-tuning of these representations for specific downstream tasks could yield further gains but would require careful regularization to avoid overfitting in small-sample regimes. Second, the noise-resolution trade-off observed in SVI super-resolution where high-frequency satellite features degrade cross-scale generalization highlights the need for adaptive feature selection mechanisms that account for the target spatial granularity. Third, while the Planetary Prediction Engine's anti-leakage protocols (Feature Gate, Split-Isolated Imputation) provide strong safeguards, formal verification of causal direction filters remains an open challenge, particularly for targets with complex, bidirectional relationships to candidate covariates. Fourth, our epidemiological nowcasting evaluation is limited to a single outbreak; validation across diverse pathogens, geographies, and surveillance infrastructures would strengthen claims of generalization.

Looking ahead, we identify several promising directions: (i) extending the framework to spatiotemporal forecasting with longer prediction horizons via temporal foundation models; (ii) scaling the Intelligent Data Selection pipeline to incorporate real-time streaming data sources (e.g., social media signals, mobility traces) for continuous model updating; and (iii) developing an ensemble architecture that combines spatial transmission models for frontier detection with mechanistic nowcasting models for resource allocation in operational outbreak response.

\section{Methods}
\label{sec:methods}

In this section, we introduce technical details of the three modular stages of the prediction workflow. The system follows rigorous modular software engineering principles: each stage operates on well-defined inputs and outputs, with no shared mutable state between stages. Data artifacts (DataFrames, GeoJSON geometries, mobility matrices) are passed between stages via opaque handles rather than serialized into LLM prompts, avoiding context-window limitations. All LLM calls use temperature  $=0$ for greater reproducibility.

\begin{figure*}[t]
    \centering
    \resizebox{\linewidth}{!}{
\begin{tikzpicture}[node distance=0.4cm and 2.8cm]

    \node[toolbox=primary] (s1_llm) {\textbf{LLM Orchestrator}\\[2pt]\color{darkgray!80}Query Interpretation \& Hypothesis Generation};
    \node[below=0.3cm of s1_llm, toolbox=primary] (s1_lock) {\textbf{Geographic Constraint Lock}\\[2pt]\color{darkgray!80}Extract spatial granularity \& join-keys};
    \node[below=0.3cm of s1_lock, toolbox=primary] (s1_signal) {\textbf{Grounded Signal Discovery}\\[2pt]\color{darkgray!80}Identify direct \& causal proxy signals};
    \node[below=0.3cm of s1_signal, toolbox=primary] (s1_fetch) {\textbf{Dynamic Data Retrieval}\\[2pt]\color{darkgray!80}Established APIs \& Live Open-Web Search};
    \node[below=0.3cm of s1_fetch, toolbox=primary] (s1_score) {\textbf{Provenance Prioritization}\\[2pt]\color{darkgray!80}5-dim scoring (5x license/provenance weight)};

    \node[above=0.6cm of s1_llm, stagetitle=primary] (s1_title) {Stage 1: Intelligent Data Selection};
    \node[below=0.1cm of s1_title, stagesubtitle] (s1_subtitle) {Dynamic Covariate Discovery Pipeline};

    \node[right=3.6cm of s1_llm, toolbox=success] (s2_llm) {\textbf{LLM Orchestrator}\\[2pt]\color{darkgray!80}Curation \& Alignment Management};
    \node[below=0.3cm of s2_llm, toolbox=success] (s2_gate) {\textbf{Feature Gate (Anti-Leakage)}\\[2pt]\color{darkgray!80}Unfair Advantage Removal (4 criteria)};
    \node[below=0.3cm of s2_gate, toolbox=success] (s2_fuse) {\textbf{Multimodal Feature Fusion}\\[2pt]\color{darkgray!80}Fuse tabular covariates + PDFM + AlphaEarth};
    \node[below=0.3cm of s2_fuse, toolbox=success] (s2_impute) {\textbf{Split-Isolated Imputation}\\[2pt]\color{darkgray!80}Compute imputation values strictly on train set};
    \node[below=0.3cm of s2_impute, toolbox=success] (s2_split) {\textbf{Geographic \& Random Splits}\\[2pt]\color{darkgray!80}Create isolated Train / Val / Test partitions};

    \node[above=0.6cm of s2_llm, stagetitle=success] (s2_title) {Stage 2: Multimodal Curation};
    \node[below=0.1cm of s2_title, stagesubtitle] (s2_subtitle) {Dataset Curation Toolkit};

    \node[right=3.6cm of s2_llm, toolbox=purple] (s3_llm) {\textbf{LLM Orchestrator}\\[2pt]\color{darkgray!80}Model Search \& Optimization Strategy};
    \node[below=0.3cm of s3_llm, toolbox=purple] (s3_task) {\textbf{Task \& Objective Alignment}\\[2pt]\color{darkgray!80}Spatial Reg, Super-Res, STM, SEIR Nowcast};
    \node[below=0.3cm of s3_task, toolbox=purple] (s3_search) {\textbf{Automated Model Search}\\[2pt]\color{darkgray!80}\texttt{linear}, \texttt{boost}, \texttt{xgboost}, \texttt{mlp} (Keras)};
    \node[below=0.3cm of s3_search, toolbox=purple] (s3_guard) {\textbf{Overfitting Guard Protocol}\\[2pt]\color{darkgray!80}Pre-train Risk Assess \& Overfit Gap ($\Delta R^2$)};
    \node[below=0.3cm of s3_guard, toolbox=purple] (s3_correct) {\textbf{Self-Correction Loop}\\[2pt]\color{darkgray!80}Detect test degradation \& restart search};

    \node[above=0.6cm of s3_llm, stagetitle=purple] (s3_title) {Stage 3: Automated Prediction};
    \node[below=0.1cm of s3_title, stagesubtitle] (s3_subtitle) {Model Search Toolkit};

    \node[above=1.5cm of s1_title, externalbox=warning, minimum width=4.8cm, xshift=-0.5cm] (query) {\textbf{Natural Language Query}\\[4pt]\textit{``Forecast obesity rates for US counties''}\\\textit{``Nowcast active Ebola dynamics''}};

    \node[left=0.6cm of query, externalbox=success, minimum width=3.2cm] (byod) {\textbf{BYOD}\\[3pt]\color{darkgray}Bring Your Own Data\\[3pt]\textit{CSVs, Shapefiles, Tables}};
    
    \node[above=1.5cm of s2_title, externalbox=primary, minimum width=5.2cm, xshift=-1.6cm] (db) {\textbf{Public \& Open-Web Repositories}\\[4pt] Data Commons, Open-Web \textbullet\ Google Earth \\Google Maps Platform Insights  };
    
    \node[above=1.5cm of s3_title, externalbox=success, minimum width=5.2cm, xshift=-1.6cm] (fm) {\textbf{Pretrained Foundation Models}\\[4pt]\textbf{PDFM:} 330/512-dim socio-demographic\\\textbf{AlphaEarth:} 64-dim satellite features};

    \begin{scope}[on background layer]
        \node[stagebox=primary, fit=(s1_title)(s1_subtitle)(s1_llm)(s1_lock)(s1_signal)(s1_fetch)(s1_score)] (stage1) {};
        \node[stagebox=success, fit=(s2_title)(s2_subtitle)(s2_llm)(s2_gate)(s2_fuse)(s2_impute)(s2_split)] (stage2) {};
        \node[stagebox=purple, fit=(s3_title)(s3_subtitle)(s3_llm)(s3_task)(s3_search)(s3_guard)(s3_correct)] (stage3) {};
    \end{scope}

    \draw[widearrow=primary] (query.south) -- (query.south |- stage1.north);

    \draw[widearrow=success] (byod.south) -- ++(0,-0.6cm) -| ([xshift=-1.5cm]stage1.north |- s1_title.north);
    
    \draw[widearrow=primary, <->] (db.south) -- ++(0,-0.6cm) -| ([xshift=1.5cm]stage1.north |- s1_title.north);
    
    \draw[widearrow=success] (fm.south) -- ++(0,-0.6cm) -| ([xshift=1.5cm]stage2.north |- s2_title.north);

    \draw[widearrow=primary] (stage1.east |- s1_signal.east) -- node[handlelabel, text width=1.8cm] {\textbf{Opaque Handles}\\[2pt]\color{darkgray}DataFrames \& GeoJSON\\[2pt]\color{primary}\textit{No shared state}} (stage2.west |- s1_signal.east);
    
    \draw[widearrow=success] (stage2.east |- s2_fuse.east) -- node[handlelabel, text width=1.8cm] {\textbf{Curated Handles}\\[2pt]\color{darkgray}Train/Val/Test Sets\\[2pt]\color{success}\textit{Opaque to LLM}} (stage3.west |- s2_fuse.east);

    \node[below=1.2cm of stage2, externalbox=purple, minimum width=8.0cm] (output) {\textbf{Trained Planetary Prediction Agent (PPA) Model \& Predictions}\\[4pt]\color{darkgray}Evaluation Metrics: $R^2$, RMSE, MAE, MAPE, Pearson $r$ \textbullet\ Returned to Client UI};
    
    \draw[widearrow=purple] (stage3.south) |- (output.east);
    \draw[dashedarrow=purple] (output.west) -| (stage1.south);

\end{tikzpicture}}
    \caption{\small The {Planetary Prediction Engine}’s multi-stage modular architecture.}
    \label{fig:method-architecture}
\end{figure*}
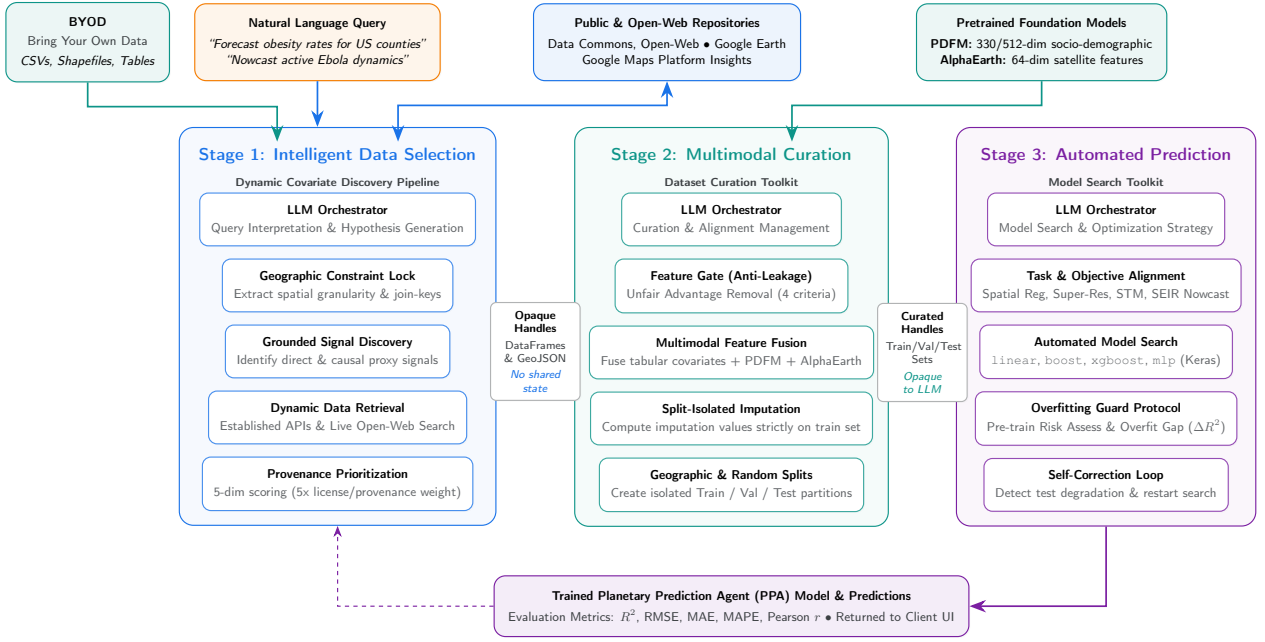

\subsection{Intelligent Data Selection}
\label{subsec:intelligent-data-selection}
A distinguishing feature of the Planetary Prediction Engine is its Intelligent Data Selection stage: a pipeline that transforms a natural-language predictive query into a prioritized, join-ready multimodal DataFrame without manual data discovery, download, or curation. Based on the geographic and temporal constraints of the user’s labeled data, the system dynamically identifies, retrieves, scores, and assembles covariates at runtime for modeling.

\subsubsection{Dynamic Covariate Discovery Pipeline }
\label{subsubsec:method-data-discrovery}
The pipeline is decomposed into six sequential sub-stages (See Appendix~\ref{sec:appendix_data_selection} for more details):

1. \textbf{Geographic Constraint Discovery:} Inspects the user-provided labeled training data to extract a geographic constraint (spatial granularity, join-key format, and temporal scope). 
2. \textbf{Grounded Signal Discovery:} Formulates domain hypotheses to draft a structured Signal Guide of candidate covariates, distinguishing between \textit{direct signals}, which are essential domain variables, and \textit{proxy signals}, variables substituting for direct signals through well-established causal relationships. Every signal is validated against published literature and official reports.  

3. \textbf{Established Repository Retrieval:} Systematically queries established, openly accessible geospatial repositories based on signal type: socio-demographic, economic, and pre-aggregated environmental statistics are retrieved from Data Commons \citep{guha2023data}; raster environmental data requiring pixel-level aggregation are processed through Google Earth Engine \citep{gorelick2017google}; and point-of-interest (POI) density metrics are obtained from the Google Maps Platform Insights.  

4. \textbf{Ad-Hoc Open-Web Discovery:} For signals not resolved by established repositories, the system performs live open-web discovery for relevant datasets (CSV, GeoJSON, Parquet, etc.) across government portals (CDC, Census, WHO, UN OCHA HDX) and academic repositories (Zenodo, Harvard Dataverse), downloading and validating them programmatically.  

5. \textbf{Provenance-First Prioritization:} Ranks datasets using a structured five-dimension scoring rubric (Provenance \& License, Spatio-Temporal Fitness, Signal Alignment, Format Quality, Redundancy). Provenance and license openness carry the highest weight (5x), ensuring the assembled covariate matrix is composed primarily of openly licensed, institutionally authoritative data.  

6. \textbf{DataFrame Assembly:} Merges all retained datasets into a single, join-ready covariate DataFrame, standardizing schemas (EPSG:4326 for geometries, ISO 8601 for dates) and generating a comprehensive data source audit and provenance metadata tracking report.

\subsubsection{Multimodal Dataset Curation with Foundation Model Embeddings}
\label{subsubsec:method-embedding-fusion}
To establish a robust, multimodal framework for geospatial modeling, the curation stage fuses the assembled tabular geospatial covariates (data sources deferred to Appendix~\ref{sec:appendix_covariates_inventory}) with pre-trained geospatial foundation models. These features represent multi-dimensional geographical properties that collectively capture the socio-demographic, physical, industrial, and ecological dimensions of target regions. The agent is therefore able to leverage pre-trained deep neural representations to encode complex, non-linear geographical and demographic characteristics. These include Population Dynamics Foundation Models (PDFM) for socio-economic latent states (330/512-dim) and AlphaEarth Foundation models for satellite imagery semantics (64-dim).

\subsection{Data preprocessing and Feature engineering}
\label{subsec:method-feature-engineering}
\subsubsection{Automated Target Leakage Mitigation (Feature Gate)}

To ensure the integrity of our empirical benchmarks and prevent unfair advantages \citep{kaufman2012leakage, kapoor2023leakage}, the PPE implements a strict \textit{Feature Gate} during dataset curation. Upon receiving the mathematical definition of the prediction target in the prompt, the agent evaluates every candidate covariate against four mandatory anti-leakage criteria:

{ \textbf{Criteria 1 (No Mathematical Components):} The covariate must not be a mathematical component, direct proxy, or sub-index used to calculate the ground truth target. }(Example: Exclude "Median Rent" if predicting "Housing Affordability Index" which is income divided by rent).

{ \textbf{Criteria 2 (No Synthetic/Shared Survey Leakage):} The covariate and target must not rely on the exact same underlying survey data or imputation models. Furthermore, when predicting population-related targets, the system restricts covariates to non-enumerative, intensive socioeconomic rates (e.g., Median Income) rather than enumerative counts (e.g., Count HousingUnit) to guarantee zero census enumeration leakage. }(Example: Exclude Census age/income demographics if predicting a synthetic ``Climate Vulnerability Score'' derived from those same Census tables). 

{ \textbf{Criteria 3 (Causal Direction Filter):} The covariate must be causally "upstream" (a driver, structural condition, or parallel confounder), not "downstream" (an effect, symptom, or response) of the target. }(Example: Exclude "Number of Delayed Deliveries" when predicting "Traffic Congestion Levels").  

\textbf{Criteria 4 (Temporal Filter):} The covariate feature must be from a time period before or during the target prediction window, preventing future information leakage.

\subsubsection{Anti-Leakage Imputation}

To ensure rigorous validation, the agent first quantifies the missingness rate for each candidate feature. Columns with insufficient coverage are discarded to mitigate overfitting to highly noisy or heavily imputed data. For the retained features, we implement a Split-Isolated Imputation protocol to prevent data leakage: imputation statistics (mean, median, or mode) are computed exclusively from the training partition and subsequently applied to the validation and test sets, thereby ensuring that downstream partitions represent strictly unseen data.

\subsubsection{Feature-Specific Engineering}

Feature engineering is applied conditionally based on the underlying data modality, which broadly comprises two categories: geospatial covariates and foundation model embeddings.

\textbf{Geospatial Covariates:} We apply standard tabular transformations to stabilize training:  
\begin{itemize}
    \item \textit{Skewness Correction}: A log(1+x) transformation is applied to features exhibiting severe skewness (|skew|> 1.0).  
    \item \textit{Outlier \& Redundancy Mitigation}:  Features are clipped to their 1st and 99th percentiles to bound extreme values, and collinearity is addressed by dropping one feature from any pair with a high Pearson correlation coefficient (|r|>0.95).  
    \item \textit{Scaling}: Variance is normalized using a standard Z-score transformation (StandardScaler). 
\end{itemize}

\textbf{Foundation Model Embeddings:} We apply L2 normalization to project location-based model embeddings onto the unit hypersphere. Crucially, we eschew independent statistical transformations (such as Z-score or logarithmic scaling) on individual dimensions to preserve the semantic topology and angular relationships of the learned representation space.

\subsection{Geospatial Prediction}
\label{subsec:geospatial-prediction}

The Automated Model Prediction stage is responsible for model training, hyperparameter optimization, and evaluation using pre-curated datasets. Operating within a strict isolation scope, the prediction agent is prohibited from fetching or mutating data, relying entirely on train and test DataFrame handles provided by the curation stage.

\subsubsection{Modeling Objectives}

The geospatial prediction stage automatically identifies the appropriate prediction task from the user's prompt and aligns the optimization objective across three core frameworks: Nowcasting, Super-Resolution, and Spatial Regression. Detailed problem formulations are explained below.

\paragraph{General Notations.}
Given a location $i$, let $X_i^{\text{cov}}$ denote automatically selected geospatial covariates, and let $X_i^{\text{emb}}$ denote location-based embedding signals generated by pretrained geospatial foundation models. The feature is constructed as $X_i = [X_i^{\text{cov}}, X_i^{\text{emb}}]$. Let $f$ parameterized by $\theta$ denote a model to train, and $y$ denote the prediction target. The general training strategy is to optimize parameter $\theta$ to $\theta^*$ which minimizes $\sum_i L(f(X_i; \theta), y_i)$, where $L(\cdot, \cdot)$ is the objective function. Using the trained model parameters $\theta^*$, we infer the prediction target at location $j$ with $\hat{y}_j = f(X_j; \theta^*)$. All three prediction tasks employ variants of this general formulation.

\paragraph{Epidemiological Nowcasting.}
Epidemiological Nowcasting refers to training on observed infection locations and predicting the infection spread frontier, i.e., the increase in infection caseloads, in a forecasting horizon (i.e., a 7-day window). Pandemic-specific signals such as infrastructure density, mobility corridors, etc., are of vital importance to prediction performance; therefore, when fetching geospatial covariates $X_i^{\text{cov}}$, the Planetary Prediction Engine automatically selects pandemic-focused signals beyond common signals (i.e., demographics). The prediction target $y_i$ in this case is $y_i^{(t)} = C_i(t + \Delta t) - C_i(t)$, where $C_i(t)$ is the total confirmed caseloads at time $t$ at location $i$, and $\Delta t$ is the nowcasting horizon (i.e., 7 days). The model optimization is conducted spatially across the total infection-risky region and temporally across all historical expanding-window folds:
$$\theta^* = \operatorname{argmin}_\theta \sum_{t} \sum_{i} L\bigl(f(X_i^{(t)}; \theta),\; y_i^{(t)}\bigr).$$

\paragraph{Super-Resolution Downscaling.}
Super-Resolution Downscaling refers to training on observed data at coarse granularity (i.e., county-level) and predicting at fine-grained granularity (i.e., ZIP-code-level) by injecting zero-label ZIP rows with geospatial features. To overcome the ecological fallacy associated with coarse regional boundaries (where uninhabited forests receive identical operational priority to high-density transit nodes), we define a spatial downscaling framework. Let $C$ represent a coarse geography (e.g., a county or health zone). Let $Z$ denote fine-resolution target localities (e.g., ZIP codes or 1\,km grid tiles). The model is trained on the coarse administrative boundaries:
$$\theta^* = \operatorname{argmin}_\theta \sum_{c \in C} L\bigl(f(X_c; \theta),\; y_c\bigr),$$
then we infer values across all fine-grained locations:
$$\hat{y}_z = f(X_z; \theta^*), \quad z \in Z.$$

\paragraph{Spatial Regression.}
Spatial Regression refers to training on observed locations and predicting missing values at the same geographic granularity. To address data sparsity and incomplete observations at a uniform geographic scale (where reporting constraints or missing surveys leave certain units without labels), we define a spatial regression framework for imputation. Let $O$ and $M$ represent observed and unobserved locations, respectively. $O$ and $M$ should be of the same granularity, i.e., both at the county-level. The model is trained on the observed locations:
$$\theta^* = \operatorname{argmin}_\theta \sum_{o \in O} L\bigl(f(X_o; \theta),\; y_o\bigr),$$
then we infer missing values across unobserved locations:
$$\hat{y}_m = f(X_m; \theta^*), \quad m \in M.$$

\subsubsection{Model Search}

\paragraph{Model Families and Hyperparameter Tuning.}
The Planetary Prediction Engine evaluates four core supervised model families (key hyperparameters listed in Table~\ref{tab:model-families}): Regularized Linear Models, Histogram-Based Gradient Boosting (Boost), Extreme Gradient Boosting (XGBoost), and Multi-Layer Perceptrons (MLP). To guard against resource exhaustion in automated search loops, all model families are bound by hard safety caps on model complexity.

\begin{table}[h]
\centering
\small
\caption{Model types supported by the Planetary Prediction Engine.}
\label{tab:model-families}
\begin{tabular}{llp{3.5cm}p{6.5cm}}
\toprule
\textbf{Model} & \textbf{Class} & \textbf{Implementation} & \textbf{Key Hyperparams} \\
\midrule
linear & Linear & Scikit-Learn (Ridge, Lasso, ElasticNet)~\citep{pedregosa2011scikit} & Regularization Type, Penalty $\alpha$, L1 ratio $\rho$ \\
\midrule
boost & HistBoost, XGBoost~\citep{chen2016xgboost} & Scikit-Learn HistGradientBoosting Regressor; xgboost.XGBRegressor & Learning rate, Max leaf nodes,Min samples per leaf, Loss Type,Max tree depth, \# of estimators,Subsample, Regularization \\
\midrule
mlp & MLPKeras & Keras & Hidden layer sizes, Dropout rate, Optimizer configs \\
\bottomrule
\end{tabular}
\end{table}

\paragraph{Model Validation and Selection Strategies.}
Model selection can be executed via sequential searches or parallelized batch search. The agent evaluates configurations using any of the following validation strategies:
\begin{itemize}
    \item \textbf{Random Split:} Divides the training data into an 80/20 train/validation split using a fixed random seed, making it ideal for standard i.i.d.\ tabular datasets and fast exploratory parameter sweeps.
    \item \textbf{Spatial Group Split:} Partitions data along geographic boundaries, which prevents spatial autocorrelation leakage and provides an accurate measure of true out-of-region generalization \citep{roberts2017cross, meyer2019importance}.
    \item \textbf{K-Fold Cross-Validation:} Implements a standard 3-fold cross-validation scheme to ensure stable performance estimates on small datasets or high-variance tasks.
\end{itemize}

\paragraph{Model Overfitting Guard Protocol.}
To ensure generalization across unseen geographies, the agent implements a multi-layered Overfitting Guard Protocol:
\begin{enumerate}
    \item \textbf{Pre-Training Risk Assessment:} Prior to training, the agent executes a heuristic check on dataset characteristics (sample size $n$, feature-to-sample ratio $p/n$, and spatial grouping) to classify overfitting risk as Low, Medium, or High. If classified as Medium, tree configurations are restricted to conservative depths. If High, strong regularization is enforced, biasing selection towards regularized linear models.
    \item \textbf{Post-Training Self-Correction Protocol:} To mimic the diagnostic judgment of a human practitioner, the agent implements a self-correction protocol: if the validation-set evaluation reveals catastrophic generalization failure (e.g., negative validation metrics or a large train-validation gap), the agent discards the current model, increases regularization constraints, and restarts with conservative configurations. The self-correction loop is constrained to a single iteration to prevent infinite optimization cycles.
\end{enumerate}

\paragraph{Evaluation Metrics.}
All models report $R^2$, Root Mean Squared Error (RMSE), Mean Absolute Error (MAE), Mean Absolute Percentage Error (MAPE), and Pearson correlation ($r$).

\subsection{Agent Execution and Runtime Profile}
As an example, in case of Ebola nowcasting, we noticed that the agent executed 793 steps across 3 sessions above:
\begin{itemize}
    \item Session 1 (15.8 min): Initialization and baseline reproduction (Steps 0–194)
    \item Session 2 (23.8 min): Feature vectorization and pipeline acceleration (Steps 195–542)
    \item Session 3 (15.6 min): Ablation evaluations and hyperparameter searches (Steps 543–792)
\end{itemize}

\section{Conclusion}
\label{sec:conclusion}
We present the Planetary Prediction Engine, an autonomous AI system that translates natural-language queries into executed geospatial predictive models, spanning spatial regression, super-resolution downscaling, spatial transmission modeling, and epidemiological nowcasting. The Planetary Prediction Engine addresses a fundamental bottleneck in planetary analytics: the labor-intensive, expertise-dependent process of discovering, curating, and fusing heterogeneous geospatial data with domain-appropriate modeling architectures.

Our empirical evaluation across a multidimensional benchmark matrix spanning the Global North and Global South, public health and environmental risk, and data-rich and data-scarce regimes demonstrates that the Planetary Prediction Engine achieves expert-level or superior performance across all evaluated paradigms. For spatial regression on 21 US CDC health indicators, the system achieves a mean $R^2$ of 76.8\%, exceeding the manually curated expert baseline of 60\%. For super-resolution food security downscaling in Nigeria, the Planetary Prediction Engine doubles baseline accuracy ($R^2$ of 66.1\% vs.\ 31.5\%). For real-time nowcasting of the 2026 DRC Ebola outbreak, the system achieves a Recall@10 of 83.3\% in predicting newly invaded health zones, representing a 10.3 percentage point absolute improvement over the public state-of-the-art.

We believe this work represents a meaningful step toward democratizing geospatial prediction, making it accessible to researchers, humanitarian organizations, and policymakers who need expert-level models but may lack the specialized engineering teams traditionally required to build them. the Planetary Prediction Engine makes predictive modeling accessible by shifting the researcher's role from manual data engineering to high-level hypothesis direction. Ultimately, it paves the way towards building geospatial prediction models useful for real-world applications in a fast, reliable and impactful manner.

\subsection*{Acknowledgements}
\label{sec:acknowledgements}

We are grateful to the UN World Food Programme (WFP) and Vulnerability Analysis and Mapping (VAM) team for the data and research support. We also thank the Institut National de Recherche Biomédicale (INRB) for their collaboration on the DRC Ebola nowcasting, and the teams behind Data Commons, Google Earth Engine, Population Dynamics Foundation Models, and AlphaEarth for providing the foundational data and model infrastructure that powers the Planetary Prediction Engine.

We extend our sincere gratitude to Aviv Slobodkin, Hamsa Subramanian, Jeremy Amez-Droz, Joydeep Paul, Lambert Rosique, Lily Mihalkova, Milind Tambe, and Tim Thelin for their insightful discussions and valuable feedback on this work.

\bibliography{references}

@article{aygun2026era,
  title={An AI system to help scientists write expert-level empirical software},
  author={Ayg{\"u}n, Eser and Belyaeva, Anastasiya and Comanici, Gheorghe and Coram, Marc and Cui, Hao and Garrison, Jake and Johnston, Renee and Kast, Anton and McLean, Cory Y and Norgaard, Peter and others},
  journal={Nature},
  pages={1--3},
  year={2026},
  publisher={Nature Publishing Group UK London}
}

@inproceedings{bran2023chemcrow,
    title={ChemCrow: Augmenting Large-Language Models with Chemistry Tools},
    author={Bran, Andres M and Cox, Sam and Schilter, Oliver and Baldassari, Carlo and White, Andrew D and Schwaller, Philippe},
    booktitle={NeurIPS 2023 Foundation Models for Decision Making Workshop},
    year={2023}
  }

@article{meyer2017spatiotemporal,
  title={Spatio-Temporal Analysis of Epidemic Phenomena Using the {R} Package surveillance},
  author={Meyer, Sebastian and Held, Leonhard and H{\"o}hle, Michael},
  journal={Journal of Statistical Software},
  volume={77},
  number={11},
  pages={1--55},
  year={2017},
  doi={10.18637/jss.v077.i11}
}

@book{cressie1993statistics,
    title={Statistics for Spatial Data},
    author={Cressie, Noel},
    year={1993},
    publisher={Wiley}
  }

@inproceedings{das2024timesfm,
    title={A Decoder-Only Foundation Model for Time-Series Forecasting},
    author={Das, Abhimanyu and Kong, Weihao and Leber, Andrew and Mathews, Rajat and Sen, Ravikumar},
    booktitle={Proceedings of the 41st International Conference on Machine Learning (ICML)},
    year={2024}
  }

@inproceedings{feurer2015efficient,
    title={Efficient and Robust Automated Machine Learning},
    author={Feurer, Matthias and Klein, Aaron and Eggensperger, Katharina and Springenberg, Jost Tobias and Blum, Manuel and Hutter, Frank},
    booktitle={Advances in Neural Information Processing Systems (NeurIPS)},
    year={2015}
  }

@inproceedings{guo2024dsagent,
    title={{DS-Agent}: Automated Data Science by Empowering Large Language Models with Case-Based Reasoning},
    author={Guo, Siyuan and Deng, Cheng and Wen, Ying and Chen, Hechang and Li, Yi and Sun, Changlong},
    booktitle={Proceedings of the 41st International Conference on Machine Learning (ICML)},
    year={2024}
  }

@article{hong2024data,
    title={Data Interpreter: An {LLM} Agent for Data Science},
    author={Hong, Sirui and Zhuge, Yizhang and Chen, Jonathan and Zheng, Xiawu and Cheng, Yu and Wang, Jinlin and Li, Jian and Wang, Zili and Lin, Dahua and Yu, Jifeng and others},
    journal={arXiv preprint arXiv:2402.18679},
    year={2024}
  }

@article{jean2016combining,
    title={Combining Satellite Imagery and Machine Learning to Predict Poverty},
    author={Jean, Neal and Burke, Marshall and Xie, Michael and Davis, W. Matthew and Lobell, David B. and Ermon, Stefano},
    journal={Science},
    volume={353},
    number={6301},
    pages={790--794},
    year={2016}
  }

@article{wu2023autogen,
    title={{AutoGen}: Enabling Next-Gen {LLM} Applications via Multi-Agent Conversation},
    author={Wu, Qingyun and Bansal, Gagan and Zhang, Jieyu and Wu, Yiran and Li, Beibin and Zhu, Erkang and Jiang, Li and Zhang, Xiaoyun and Zhang, Shaokun and Liu, Jiale and others},
    journal={arXiv preprint arXiv:2308.08155},
    year={2023}
  }

@article{yeh2020using,
    title={Using Publicly Available Satellite Imagery and Deep Learning to Understand Economic Well-being in {Africa}},
    author={Yeh, Christopher and Perez, Anthony and Driscoll, Anne and Azzari, George and Tang, Zhongyi and Lobell, David and Ermon, Stefano and Burke, Marshall},
    journal={Nature Communications},
    volume={11},
    number={1},
    pages={2583},
    year={2020}
  }

@article{wei2022chain,
  title={Chain-of-Thought Prompting Elicits Reasoning in Large Language Models},
  author={Wei, Jason and Wang, Xuezhi and Schuurmans, Dale and Bosma, Maarten and Ichter, Brian and Xia, Fei and Chi, Ed and Le, Quoc V and Zhou, Denny},
  journal={Advances in Neural Information Processing Systems},
  volume={35},
  pages={24824--24837},
  year={2022}
}

@article{brown2024large,
  title={Large Language Monkeys: Scaling Inference Compute with Repeated Sampling},
  author={Brown, Bradley and Juravsky, Jordan and Ehrlich, Ryan and Clark, Ronald and Le, Quoc V and R{\'e}, Christopher and Mirhoseini, Azalia},
  journal={arXiv preprint arXiv:2407.21787},
  year={2024}
}

@article{madaan2023selfrefine,
  title={Self-Refine: Iterative Refinement with Self-Feedback},
  author={Madaan, Aman and Tandon, Niket and Gupta, Prakhar and Hallinan, Skyler and Gao, Luyu and Wiegreffe, Sarah and Alon, Uri and Dziri, Nouha and Prabhumoye, Shachi and Yang, Yiming and others},
  journal={Advances in Neural Information Processing Systems},
  volume={36},
  year={2023}
}

@article{snell2024scaling,
  title={Scaling {LLM} Test-Time Compute Optimally can be More Effective than Scaling Model Parameters},
  author={Snell, Charlie and Lee, Jaehoon and Xu, Kelvin and Kumar, Aviral},
  journal={arXiv preprint arXiv:2408.03314},
  year={2024}
}

@article{chen2023frugalgpt,
  title={{FrugalGPT}: How to Use Large Language Models While Reducing Cost and Improving Performance},
  author={Chen, Lingjiao and Zaharia, Matei and Zou, James},
  journal={arXiv preprint arXiv:2305.05176},
  year={2023}
}

@article{ong2024routellm,
  title={{RouteLLM}: Learning to Route {LLMs} with Preference Data},
  author={Ong, Isaac and Almahairi, Amjad and Wu, Vincent and Chiang, Wei-Lin and Wu, Tianhao and Gonzalez, Joseph E and Kadous, M Waleed and Stoica, Ion},
  journal={arXiv preprint arXiv:2406.18665},
  year={2024}
}

@article{yao2023react,
  title={{ReAct}: Synergizing Reasoning and Acting in Language Models},
  author={Yao, Shunyu and Zhao, Jeffrey and Yu, Dian and Du, Nan and Shafran, Izhak and Narasimhan, Karthik and Cao, Yuan},
  journal={International Conference on Learning Representations},
  year={2023}
}

@article{schick2023toolformer,
  title={Toolformer: Language Models Can Teach Themselves to Use Tools},
  author={Schick, Timo and Dwivedi-Yu, Jane and Dess{\`i}, Roberto and Raileanu, Roberta and Lomeli, Maria and Hambro, Eric and Zettlemoyer, Luke and Cancedda, Nicola and Scialom, Thomas},
  journal={Advances in Neural Information Processing Systems},
  volume={36},
  year={2023}
}

@article{wang2023planandsolve,
  title={Plan-and-Solve Prompting: Improving Zero-Shot Chain-of-Thought Reasoning by Large Language Models},
  author={Wang, Lei and Xu, Wanyu and Lan, Yihuai and Hu, Zhiqiang and Lan, Yunshi and Lee, Roy Ka-Wei and Lim, Ee-Peng},
  journal={Proceedings of the 61st Annual Meeting of the Association for Computational Linguistics},
  year={2023}
}

@article{wang2024mixture,
  title={Mixture-of-Agents Enhances Large Language Model Capabilities},
  author={Wang, Junlin and Wang, Jue and Athiwaratkun, Ben and Zhang, Ce and Zou, James},
  journal={arXiv preprint arXiv:2406.04692},
  year={2024}
}

@article{du2024improving,
  title={Improving Factuality and Reasoning in Language Models through Multiagent Debate},
  author={Du, Yilun and Li, Shuang and Torralba, Antonio and Tenenbaum, Joshua B and Mordatch, Igor},
  journal={Proceedings of the 41st International Conference on Machine Learning},
  year={2024}
}

@article{zaharia2024shift,
  title={The Shift from Models to Compound {AI} Systems},
  author={Zaharia, Matei and Khattab, Omar and Chen, Lingjiao and Gonzalez, Joseph E and Stoica, Ion and others},
  journal={Berkeley AI Research Blog},
  year={2024}
}

@article{wang2023selfconsistency,
  title={Self-Consistency Improves Chain of Thought Reasoning in Language Models},
  author={Wang, Xuezhi and Wei, Jason and Schuurmans, Dale and Le, Quoc and Chi, Ed and Narang, Sharan and Chowdhery, Aakanksha and Zhou, Denny},
  journal={International Conference on Learning Representations},
  year={2023}
}

@article{yao2023tree,
  title={Tree of Thoughts: Deliberate Problem Solving with Large Language Models},
  author={Yao, Shunyu and Yu, Dian and Zhao, Jeffrey and Shafran, Izhak and Griffiths, Thomas L and Cao, Yuan and Narasimhan, Karthik},
  journal={Advances in Neural Information Processing Systems},
  volume={36},
  year={2023}
}

@inproceedings{krechetova2025geobenchx,
  title={GeoBenchX: Benchmarking LLMs in agent solving multistep geospatial tasks},
  author={Krechetova, Varvara and Kochedykov, Denis},
  booktitle={Proceedings of the 1st ACM SIGSPATIAL International Workshop on Generative and Agentic AI for Multi-Modality Space-Time Intelligence},
  pages={27--35},
  year={2025}
}

@article{mialon2023gaia,
  title={{GAIA}: A Benchmark for General {AI} Assistants},
  author={Mialon, Gr{\'e}goire and Fourrier, Cl{\'e}mentine and Swift, Craig and Wolf, Thomas and LeCun, Yann and Scialom, Thomas},
  journal={arXiv preprint arXiv:2311.12983},
  year={2023}
}

@article{yang2018hotpotqa,
  title={{HotpotQA}: A Dataset for Diverse, Explainable Multi-hop Question Answering},
  author={Yang, Zhilin and Qi, Peng and Zhang, Saizheng and Bengio, Yoshua and Cohen, William W and Salakhutdinov, Ruslan and Manning, Christopher D},
  journal={Proceedings of the Conference on Empirical Methods in Natural Language Processing},
  year={2018}
}

@article{kim2025scaling,
  title={Towards a Science of Scaling Agent Systems},
  author={Kim, Yubin and Gu, Ken and Park, Chanwoo and Park, Chunjong and Schmidgall, Samuel and Heydari, A. Ali and Yan, Yao and Zhang, Zhihan and Zhuang, Yuchen and Malhotra, Mark and Liang, Paul Pu and Park, Hae Won and Yang, Yuzhe and Xu, Xuhai and Du, Yilun and Patel, Shwetak and Althoff, Tim and McDuff, Daniel and Liu, Xin},
  journal={arXiv preprint arXiv:2512.08296},
  year={2025}
}

@article{trivedi2022musique,
  title={{MuSiQue}: Multihop Questions via Single Hop Question Composition},
  author={Trivedi, Harsh and Balasubramanian, Niranjan and Khot, Tushar and Sabharwal, Ashish},
  journal={Transactions of the Association for Computational Linguistics},
  volume={10},
  pages={539--554},
  year={2022}
}

@misc{clay2024foundation,
  title={{Clay}: An Open Source Foundation Model for Earth Observation},
  author={{Clay Foundation}},
  year={2024},
  howpublished={\url{https://huggingface.co/made-with-clay/Clay}},
  note={Accessed: 2026-08-01}
}

@inproceedings{jakubik2023prithvi,
  title={Foundation Models for Generalist Geospatial Artificial Intelligence},
  author={Jakubik, Johannes and Muszynski, Michal and Moffeit, Chris and Gao, Rohit and Ramasubramanian, Sujit and others},
  booktitle={Advances in Neural Information Processing Systems (NeurIPS)},
  year={2023}
}

@article{brown2025alphaearth,
  title={{AlphaEarth} Foundations: An Embedding Field Model for Accurate and Efficient Global Mapping from Sparse Label Data},
  author={Brown, Christopher F. and Kazmierski, Michal R. and Pasquarella, Valerie J. and others},
  journal={arXiv preprint arXiv:2507.22291},
  year={2025}
}

@article{lu2024aiscientist,
  title={The {AI} Scientist: Towards Fully Automated Open-Ended Scientific Discovery},
  author={Lu, Chris and Lu, Cong and Lange, Robert Tjarko and Foerster, Jakob and Clune, Jeff and Ha, David},
  journal={arXiv preprint arXiv:2408.06292},
  year={2024}
}

@article{real2020automl,
  title={{AutoML-Zero}: Evolving Machine Learning Algorithms from Scratch},
  author={Real, Esteban and Liang, Chen and So, David and Le, Quoc},
  journal={International Conference on Machine Learning (ICML)},
  year={2020}
}

@article{novikov2025alphaevolve,
  title={{AlphaEvolve}: A Coding Agent for Scientific and Algorithmic Discovery},
  author={Novikov, Alexander and V{\~u}, Ng{\^a}n and Eisenberger, Marvin and Dupont, Emilien and Huang, Po-Sen and Wagner, Adam Zsolt and Shirobokov, Sergey and Kozlovskii, Borislav and Ruiz, Francisco J. R. and Mehrabian, Abbas and Kumar, M. Pawan and See, Abigail and Chaudhuri, Swarat and Holland, George and Davies, Alex and Nowozin, Sebastian and Kohli, Pushmeet and Balog, Matej},
  journal={arXiv preprint arXiv:2506.13131},
  year={2025}
}

@article{hoehle2017nowcasting,
  title={Epidemiological Nowcasting: Assessing the Progress of Outbreaks in Real Time},
  author={H{\"o}hle, Michael},
  journal={Epidemiology and Infection},
  volume={145},
  number={15},
  pages={3100--3110},
  year={2017}
}

@article{mcgough2020nowcasting,
  title={Nowcasting by {Bayesian} Smoothing: A Framework for Real-Time Outbreak Tracking},
  author={McGough, Sarah F and Johansson, Michael A and Lipsitch, Marc and Santillana, Mauricio},
  journal={PLOS Computational Biology},
  volume={16},
  number={4},
  pages={e1007741},
  year={2020}
}

@article{bastos2019modelling,
  title={A Modelling Framework for Nowcasting Disease Incidence with Applications to Malaria and Severe Acute Respiratory Infection},
  author={Bastos, Leonardo S and Economou, Theodoros and Gomes, Marcelo FC and Villela, Daniel AM and Coelho, Flavio C and Cruz, Oswaldo G and Carvalho, Marilia S and Code{\c{c}}o, Claudia T},
  journal={Statistics in Medicine},
  volume={38},
  number={24},
  pages={4853--4867},
  year={2019}
}

@article{park2004mrp,
  title={Bayesian Multilevel Estimation with Poststratification: State-Level Estimates from National Polls},
  author={Park, David K. and Gelman, Andrew and Bafumi, Joseph},
  journal={Political Analysis},
  volume={12},
  number={4},
  pages={375--385},
  year={2004},
  publisher={Cambridge University Press}
}

@article{funk2015climate,
  title={The Climate Hazards Group InfraRed Precipitation with Station Data ({CHIRPS}): A New Dataset for Monitoring Extremes},
  author={Funk, Chris and Peterson, Pete and Landsfeld, Martin and Pedreros, Diego and Verdin, James and Shukla, Shraddhanand and Husak, Gregory and Rowland, James and Harrison, Laura and Hoell, Andrew and others},
  journal={Scientific Data},
  volume={2},
  number={1},
  pages={150066},
  year={2015}
}

@article{burke2021satellite,
  title={Using Satellite Imagery to Understand and Promote Sustainable Development},
  author={Burke, Marshall and Driscoll, Anne and Lobell, David B. and Ermon, Stefano},
  journal={Science},
  volume={371},
  number={6535},
  pages={eabe8628},
  year={2021}
}

@article{simini2012universal,
  title={A Universal Model for Mobility and Migration Patterns},
  author={Simini, Filippo and Gonz{\'a}lez, Marta C and Maritan, Amos and Barab{\'a}si, Albert-L{\'a}szl{\'o}},
  journal={Nature},
  volume={484},
  number={7392},
  pages={96--110},
  year={2012}
}

@article{wesolowski2012quantifying,
  title={Quantifying the Impact of Human Mobility on Malaria Transmission in {Kenya}},
  author={Wesolowski, Amy and Eagle, Nathan and Tatem, Andrew J and Smith, David L and Noor, Abdisalan M and Snow, Robert W and Buckee, Caroline O},
  journal={Science},
  volume={338},
  number={6104},
  pages={267--270},
  year={2012}
}

@article{roberts2017cross,
  title={Cross-Validation Strategies for Data with Temporal, Spatial, Hierarchical or Phylogenetic Structure},
  author={Roberts, David R and Bahn, Volker and Ciuti, Simone and Boyce, Mark S and Elith, Jane and Guillera-Arroita, Gurutzeta and Hauenstein, Severin and El-Gabbas, Ahmed and Rai{\ss}, J{\"o}rg and Dormann, Carsten F},
  journal={Ecography},
  volume={40},
  number={8},
  pages={913--929},
  year={2017}
}

@article{meyer2019importance,
  title={Importance of Spatial Predictor Variable Selection in Machine Learning Applications - Moving from Local to Spatial Cross-Validation},
  author={Meyer, Hanna and Reudenbach, Christoph and Hengl, Tomislav and Katurji, Marwan and Nauss, Thomas},
  journal={Ecological Modelling},
  volume={411},
  pages={108815},
  year={2019}
}

@inproceedings{wang2021flaml,
  title={{FLAML}: A Fast and Lightweight Automated Machine Learning Library},
  author={Wang, Chi and Wu, Qingyun and Weimer, Markus and Zhu, Erkang},
  booktitle={Proceedings of Machine Learning and Systems (MLSys)},
  volume={3},
  pages={434--447},
  year={2021}
}

@article{flanagan2011social,
  title={A Social Vulnerability Index for Disaster Management},
  author={Flanagan, Barry E and Gregory, Edward W and Hallisey, Elaine J and Heitgerd, Janet L and Lewis, Brian},
  journal={Journal of Homeland Security and Emergency Management},
  volume={8},
  number={1},
  pages={0000102202154151000271},
  year={2011}
}

@techreport{ipc2021technical,
  title={{Integrated Food Security Phase Classification Technical Manual Version 3.1}: Evidence and Standards for Better Food Security and Nutrition Decisions},
  author={{IPC Global Partners}},
  institution={Food and Agriculture Organization of the United Nations (FAO)},
  address={Rome, Italy},
  year={2021}
}

@book{openshaw1984modifiable,
  title={The Modifiable Areal Unit Problem},
  author={Openshaw, Stan},
  year={1984},
  publisher={Geo Books Norwich}
}

@techreport{wfp2021hungermap,
  title={{HungerMap LIVE}: Near Real-Time Food Security Monitoring Systems},
  author={{World Food Programme}},
  institution={WFP Hunger Monitoring Unit},
  address={Rome, Italy},
  year={2021}
}

@book{kalkuhl2016food,
  title={Food Price Volatility and Its Implications for Food Security and Policy},
  author={Kalkuhl, Matthias and von Braun, Joachim and Torero, Maximo},
  year={2016},
  publisher={Springer Nature}
}

@inproceedings{luxen2011real,
  title={Real-Time Routing with {OpenStreetMap} Data},
  author={Luxen, Dennis and Vetter, Christian},
  booktitle={Proceedings of the 19th ACM SIGSPATIAL International Conference on Advances in Geographic Information Systems},
  pages={513--516},
  year={2011}
}

@article{haklay2008openstreetmap,
  title={{OpenStreetMap}: User-Generated Street Maps},
  author={Haklay, Mordechai and Weber, Patrick},
  journal={IEEE Pervasive Computing},
  volume={7},
  number={4},
  pages={12--18},
  year={2008}
}

@article{atkinson2013downscaling,
  title={Downscaling in Remote Sensing},
  author={Atkinson, Peter M},
  journal={International Journal of Applied Earth Observation and Geoinformation},
  volume={22},
  pages={106--114},
  year={2013}
}

@article{gorelick2017google,
  title={{Google Earth Engine}: Planetary-Scale Geospatial Analysis for Everyone},
  author={Gorelick, Noel and Hancher, Matt and Dixon, Mike and Ilyushchenko, Simon and Thau, David and Moore, Rebecca},
  journal={Remote Sensing of Environment},
  volume={202},
  pages={18--27},
  year={2017}
}

@inproceedings{guha2023data,
  title={{Data Commons}: Organising the World's Public Data},
  author={Guha, Ramanathan and Alon, Noga and Kaluza, Bojan and Ramakrishnan, Raghu},
  booktitle={Proceedings of the ACM Web Conference 2023},
  pages={3900--3908},
  year={2023}
}

@article{kaufman2012leakage,
  title={Leakage in Data Mining: Formulation, Detection, and Avoidance},
  author={Kaufman, Shachar and Rosset, Saharon and Perlich, Claudia and Stitelman, Ori},
  journal={ACM Transactions on Knowledge Discovery from Data (TKDD)},
  volume={6},
  number={4},
  pages={1--21},
  year={2012}
}

@article{kapoor2023leakage,
  title={Leakage and the Reproducibility Crisis in Machine-Learning-Based Science},
  author={Kapoor, Sayash and Narayanan, Arvind},
  journal={Patterns},
  volume={4},
  number={9},
  pages={100804},
  year={2023}
}

@inproceedings{chen2016xgboost,
  title={{XGBoost}: A Scalable Tree Boosting System},
  author={Chen, Tianqi and Guestrin, Carlos},
  booktitle={Proceedings of the 22nd ACM SIGKDD International Conference on Knowledge Discovery and Data Mining},
  pages={785--794},
  year={2016}
}

@article{pedregosa2011scikit,
  title={{Scikit-learn}: Machine Learning in Python},
  author={Pedregosa, Fabian and Varoquaux, Ga{\"e}l and Gramfort, Alexandre and Michel, Vincent and Thirion, Bertrand and Grisel, Olivier and Blondel, Mathieu and Prettenhofer, Peter and Weiss, Ron and Dubourg, Vincent and others},
  journal={Journal of Machine Learning Research},
  volume={12},
  pages={2825--2830},
  year={2011}
}

@misc{who2026ebola,
  title={Ebola Disease Caused by {Bundibugyo} Virus -- {Democratic Republic of the Congo}},
  author={{World Health Organization}},
  year={2026},
  howpublished={\url{https://www.who.int/emergencies/disease-outbreak-news/item/2026-DON614}},
  note={Disease Outbreak News, 22 May 2026. Accessed: 2026-08-01}
}

@inproceedings{hong2023metagpt,
  title={{MetaGPT}: Meta Programming for A Multi-Agent Collaborative Framework},
  author={Hong, Sirui and Zheng, Xiawu and Chen, Jonathan and Cheng, Yu and Zhang, Ceyao and Wang, Zili and Yau, Steven Ka Shing and Lin, Dahua and Zhou, Liyang and Ran, Chenyu and others},
  booktitle={International Conference on Learning Representations (ICLR)},
  year={2024}
}

@inproceedings{jimenez2024swebench,
  title={{SWE-bench}: Can Language Models Resolve Real-World GitHub Issues?},
  author={Jimenez, Carlos E and Yang, John and Wettig, Alexander and Yang, Shunyu and Yao, Shunyu and Yang, Kexin and Narasimhan, Karthik and Yao, Shunyu},
  booktitle={International Conference on Learning Representations (ICLR)},
  year={2024}
}

@inproceedings{knyazev2024satclip,
  title={{SatCLIP}: Global General-Purpose Location Embeddings},
  author={Knyazev, Boris and Harm de Vries and Cisse, Moustapha and Courville, Aaron and Taylor, Graham W},
  booktitle={International Conference on Learning Representations (ICLR)},
  year={2024}
}

@inproceedings{tseng2023lightweight,
  title={Lightweight Foundation Model for Earth Observation},
  author={Tseng, Gabriel and Zvonkov, Ivan and Kerner, Hannah R and Rolnick, David},
  booktitle={Advances in Neural Information Processing Systems (NeurIPS)},
  year={2023}
}

@article{greenlund2022places,
  title={{PLACES}: Local Data for Better Health},
  author={Greenlund, Kurt J and Lu, Hua and Zhang, Xingyou and Holt, JB and Matthews, KA and Croft, JB},
  journal={Journal of Public Health Management and Practice},
  volume={28},
  number={S2},
  pages={S123--S130},
  year={2022}
}

@misc{cdc2020svi,
  title={{CDC/ATSDR} Social Vulnerability Index},
  author={{Centers for Disease Control and Prevention}},
  year={2020},
  publisher={Agency for Toxic Substances and Disease Registry},
  howpublished={\url{https://www.atsdr.cdc.gov/placeandhealth/svi/index.html}}
}

@techreport{fema2023nri,
  title={The National Risk Index: Technical Documentation},
  author={{Federal Emergency Management Agency}},
  year={2023},
  institution={FEMA, US Department of Homeland Security},
  address={Washington, DC}
}

@article{elvidge2017viirs,
  title={{VIIRS} night-time lights},
  author={Elvidge, Christopher D and Baugh, Kimberly and Zhizhin, Mikhail and Hsu, Feng Chi and Ghosh, Tilottama},
  journal={International Journal of Remote Sensing},
  volume={38},
  number={21},
  pages={5860--5879},
  year={2017}
}

@misc{wfp2024foodprices,
  title={{Global Food Prices Database}},
  author={{World Food Programme, Vulnerability Analysis and Mapping (VAM)}},
  year={2024},
  howpublished={\url{https://data.humdata.org/dataset/global-wfp-food-prices}},
  note={Accessed via HDX. Accessed: 2026-08-01}
}

@misc{inrb2026ebola,
  title={{DRC Ebola Virus Disease (BDBV) 2026 Outbreak Caseload Registry}},
  author={{Institut National de Recherche Biom{\'e}dicale (INRB)}},
  year={2026},
  howpublished={\url{https://github.com/INRB-UMIE/Ebola_DRC_2026}},
  note={Accessed: 2026-08-01}
}

@article{hersbach2020era5,
  title={The {ERA5} Global Reanalysis},
  author={Hersbach, Hans and Bell, Bill and Berrisford, Paul and Hirahara, Shoji and Hor{\'a}nyi, Andr{\'a}s and Mu{\~n}oz-Sabater, Joaqu{\'\i}n and Nicolas, Julien and Peubey, Carole and Radu, Raluca and Schepers, Dinand and others},
  journal={Quarterly Journal of the Royal Meteorological Society},
  volume={146},
  number={730},
  pages={1999--2049},
  year={2020},
  publisher={Wiley Online Library}
}

@article{tatem2017worldpop,
  title={{WorldPop}, open data for spatial demography},
  author={Tatem, Andrew J},
  journal={Scientific Data},
  volume={4},
  number={1},
  pages={170004},
  year={2017},
  publisher={Nature Publishing Group}
}

@misc{epidemiological2026bundibugyo,
  title={Real-Time Spatiotemporal Risk Modelling of the {Bundibugyo} Ebola Virus Outbreak 2026},
  author={{Epidemiological.org Consortium}},
  year={2026},
  howpublished={\url{https://www.epidemiological.org/t/real-time-spatiotemporal-risk-modelling-of-the-bundibugyo-ebola-virus-outbreak-2026/16}},
  note={Accessed: 2026-08-01}
}

@article{mbulayi2026realtime,
  title={Real-Time Epidemic Intelligence in a Public Health Emergency: The 2026 {Bundibugyo} Virus Outbreak},
  author={Mbulayi, Onesime and Akilimali, Pierre and Judge, Ciara and Gutierrez, Bernardo and Mulu, Patrick and Ibolobolo, C M and Sibo, J-C and Nkwele wa Nkwele, R and Hermann, M M and Lawanga Ontshick, L and Mukadi, D and Kanku, B and Katanga, E and {le Polain de Waroux}, Olivier and others},
  journal={The Lancet Infectious Diseases},
  year={2026},
  doi={10.1016/S1473-3099(26)00330-0},
  url={https://doi.org/10.1016/S1473-3099(26)00330-0}
}

@misc{flowminder2026population,
  author = {{Flowminder Foundation}},
  title = {Population movements from Bunia, Mongbwalu and Rwampara based on privacy secure analysis of mobile operator data from Vodacom Congo},
  year = {2026},
  month = {may},
  url = {https://www.flowminder.org/media/eagbkohv/ebola-report-update-4-june_final.pdf},
  note = {Accessed: 2026-06-03}
}

@article{bell2025earthai,
  title={Earth AI: unlocking geospatial insights with foundation models and cross-modal reasoning},
  author={Bell, Aaron and Aides, Amit and Helmy, Amr and Muslim, Arbaaz and Barzilai, Aviad and Slobodkin, Aviv and Jaber, Bolous and Schottlander, David and Leifman, George and Paul, Joydeep and others},
  journal={arXiv preprint arXiv:2510.18318},
  year={2025}
}

@techreport{wiesmann2009validation,
  title={Validation of the {World Food Programme's} Food Consumption Score and Alternative Indicators of Household Food Security},
  author={Wiesmann, Doris and Bassett, Lucy and Benson, Todd and Hoddinott, John},
  year={2009},
  number={00870},
  institution={International Food Policy Research Institute (IFPRI)},
  type={IFPRI Discussion Paper}
}

@article{didan2015modis,
  title={{MOD13A3 MODIS/Terra} Vegetation Indices Monthly {L3} Global 1km {SIN} Grid {V006}},
  author={Didan, Kamel},
  journal={NASA EOSDIS Land Processes DAAC},
  year={2015},
  doi={10.5067/MODIS/MOD13A3.006}
}

@article{agarwal2024general,
  title={General geospatial inference with a population dynamics foundation model},
  author={Agarwal, Mohit and Sun, Mimi and Kamath, Chaitanya and Muslim, Arbaaz and Sarker, Prithul and Paul, Joydeep and Yee, Hector and Sieniek, Marcin and Jablonski, Kim and Vispute, Swapnil and others},
  journal={arXiv preprint arXiv:2411.07207},
  year={2024}
}

\newpage

\appendix

\section{Benchmarks}
\label{sec:appendix_benchmarks}

\begin{itemize}
\item \textbf{SVI (Social Vulnerability Index):} County-level and ZIP-code level socioeconomic, household, minority, and housing/transportation vulnerability metrics across the United States \citep{flanagan2011social, cdc2020svi}.  
\item
  \textbf{CDC Health Variables}: 21 chronic disease and health indicator
  variables at the census tract level (e.g., Obesity, Diabetes, Stroke,
  Asthma, COPD, High Blood Pressure) from the CDC PLACES dataset
  \citep{greenlund2022places}.

\item
  \textbf{FEMA Environmental Risk}: Census-tract-level environmental and climate
  risk indices from the FEMA National Risk Index \citep{fema2023nri}.

\item
  \textbf{Nigeria Food Security Downscaling}: Predicting Food Consumption Group
  (FCG \citealp{wiesmann2009validation}) scores (continuous food consumption index) and acute food
  insecurity across Nigeria. Features include World Bank food price
  indices (OHLC food price index, inflation food price index), HungerMap
  acute food insecurity metrics \citep{wfp2021hungermap}, HDX precipitation/rainfall
  time series, HDX NDVI vegetative indices, fatality datasets, and
  Gemini processed news data. Training at ADM1 state level (N=30),
  evaluation at ADM2 LGA level (N=581).

\item \textbf{2026 DRC Ebola Outbreak:} Simulating the May–June 2026 outbreak of Ebola Bundibugyo virus (BDBV) across the Democratic Republic of Congo using official WHO surveillance reports \citep{who2026ebola, inrb2026ebola}. Datasets include the official INRB live caseload registry (\href{https://github.com/INRB-UMIE/Ebola_DRC_2026}{INRB GitHub}), HDX COD Admin 3 subnational boundaries (\href{https://data.humdata.org/dataset/cod-ab-cod}{HDX COD}), OpenStreetMap road network vectors and structural building footprints (\href{https://data.humdata.org/dataset/drc-uganda-ebola-outbreak-osm-may-2026}{HDX OSM}), Data Commons demographics, and Earth Engine environmental layers (WorldPop, ERA5-Land temperature/precipitation, Copernicus built-up density).

\item
  Ebola Outbreak Hotspot Prediction: Simulating the May--July 2026
  outbreak of Ebola Bundibugyo virus (BDBV) across Ituri, Nord-Kivu, and
  Haut-Uele provinces in the Democratic Republic of Congo using official
  WHO surveillance reports~\citep{who2026ebola} and INRB registry data~\citep{inrb2026ebola}.

  \begin{itemize}
  \item
    \textbf{Signals and data sources.} The analysis drew on three
    categories of data, all harmonised to 519 Ministry of Health
    \emph{zones de santé} (health zones) in the DRC. All
    non-confidential data sources are publicly available through an
    open-access GitHub repository.

    \begin{itemize}
    \item
      \emph{Epidemiological:} Daily confirmed BDBV case counts by
      symptom onset date from the DHIS2 linelist (INRB/INSP),
      supplemented by daily case, death, and contact-tracing indicators
      manually transcribed from the INSP Situation Report (SitRep) MVE
      PDF series. Three health zones (Oicha, Makiso-Kisangani, Lubunga)
      absent from the linelist were manually added based on SitRep
      confirmation.
    \item
      \emph{Demographic and socioeconomic:} Gridded population counts
      and density (WorldPop), socioeconomic deprivation and inequality
      indices (Climate-Conflict Vulnerability Index, CCVI), GDP per
      capita (Kummu et al.), and health facility counts and densities
      (GRID3 COD Health Facilities v8.0).
    \item
      \emph{Flowminder Mobility:} Mobile-phone-based relocation estimates from the
      Flowminder Foundation~\cite{flowminder2026population}, capturing subscriber proportions from the
      Ituri epicentre (Bunia, Mongbwalu, Rwampara) detected in other
      health zones over successive weekly windows; zone-to-zone road
      travel times via the OpenStreetMap OSRM API; and fitted gravity
      and radiation models for zone pairs not covered by Flowminder.
    \end{itemize}
  \item
    \textbf{Data preprocessing.} The raw data streams were processed
    through the following steps before entering the spatiotemporal
    invasion model:

    \begin{itemize}
    \item
      \emph{Spatial aggregation:} All raster and point-source covariates
      were resampled and aggregated to the health zone level using the
      DART pipeline. Covariates were standardised (z-scored) prior to
      model fitting; population counts were additionally
      log-transformed.
    \item
      \emph{Epidemiological signal processing:} Missing symptom onset
      dates were imputed from sample collection dates using an estimated
      delay distribution. Case time series were then nowcasted to adjust
      for reporting delays. The nowcasted incidence in each source zone
      was convolved with a discretised Gamma generation-time
      distribution, evaluated at three parameterisations with means of
      12.0, 15.3, and 18.0 days, and binned to weekly intervals to
      match the one- and two-week forecast horizons.
    \item
      \emph{Mobility matrix construction:} Flowminder empirical mobility
      estimates and parametric gravity/radiation models were combined
      into a composite mobility matrix W, which was multiplied by the
      generation-time-weighted incidence to compute the importation
      pressure into each at-risk zone.
    \item
      \emph{Vulnerability scoring:} A rank-based composite of
      health-site density, health-site count, travel time to the nearest
      facility, and CCVI deprivation was multiplied by the predicted
      invasion probability to yield a priority score rescaled to {[}0,
      1{]}.
    \end{itemize}
  \end{itemize}
\end{itemize}

%
%
%
%
%

\section{Geospatial Covariates and Foundation Models}
\label{sec:appendix_covariates_inventory}

Details of the inventories of Geospatial Covariates and Geospatial Foundation Models are listed in Table~\ref{tab:geospatial-covariates-inventory} and ~\ref{tab:foundation-models-inventory}.

\begin{table}[h]
\centering
\small
\caption{Inventory of Geospatial Covariates}
\label{tab:geospatial-covariates-inventory}
\begin{tabular}{p{3cm}p{3cm}p{6cm}p{3.5cm}}
\toprule
\textbf{Covariate Category} & \textbf{Data Source / Provider} & \textbf{Key Variables \& Dimensions Captured} & \textbf{Access Class} \\
\midrule
Socio-Demographic \& Economic & Data Commons (Google Open Source) & Population density, household income trends, poverty rates, age/gender distributions, chronic disease rates. & Public (Open API \& Free) \\
Ecological \& Environmental & Google Earth Engine (EE) & Multi-spectral imagery (Sentinel-2, Landsat, MODIS), NDVI (vegetation), soil moisture, LST anomalies. & Public (Free for Research Use) \\
Infrastructure \& Mobility & Google Maps Platform Insights & Aggregated Statistics on POIs & Public (GCP Commercial Billing) \\
Infrastructure \& Mobility & OpenStreetMap (OSM) (via OCHA HDX) & Road network vectors, administrative boundary polygons, structural building footprints. & Public (Open Source / Free) \\
Meteorological & Open-Meteo / NOAA & Daily temperature, cumulative precipitation, relative humidity, wind velocity profiles. & Public (Open API / Free) \\
\bottomrule
\end{tabular}
\end{table}

\begin{table}[h]
\centering
\small
\caption{Inventory of Geospatial Foundation Models}
\label{tab:foundation-models-inventory}
\begin{tabular}{p{4cm}p{5.5cm}p{5.5cm}}
\toprule
\textbf{Foundation Model} & \textbf{Population Dynamics Foundation Model (PDFM)} & \textbf{AlphaEarth Foundation Model} \\
\midrule
Modality / Coverage & Socio-economic \& Demographic (State, County, ZCTA, City) & Ecological \& Remote Sensing (High-resolution satellite raster) \\
Dimension & 330 / 512 & 64 \\
Key Semantic Characteristics & Encodes latent population density, mobility, economic growth, and regional demographics. & Encodes multi-scale topography, vegetation canopy, land-use semantics, and terrain profiles. \\
Access Class & Pre-General Availability + Research Use & Publicly available \\
\bottomrule
\end{tabular}
\end{table}

\section{Intelligent Data Selection Methodology \&
Rubrics}
\label{sec:appendix_data_selection}

Table~\ref{tab:dataset-prioritization-rubric} details the five-dimension scoring rubric utilized by the
Provenance-First Prioritization sub-stage to rank candidate datasets.
Provenance and license compliance carry the highest weight (5x),
ensuring the system preferentially assembles covariates from openly
licensed, institutionally authoritative sources.

\begin{table}[h]
\centering
\small
\caption{\small Five-Dimension Scoring Rubric}
\label{tab:dataset-prioritization-rubric}
\begin{tabular}{p{3.5cm}cp{8.5cm}}
\toprule
\textbf{Dimension} & \textbf{Weight} & \textbf{Criteria \& Point Allocations} \\
\midrule
Provenance \& License & High (x5) & Institutional backing and license openness. Openly licensed data from established repositories (4 pts) → government/NGO open data (3 pts) → attribution-required academic data (2 pts) → restrictive or unknown license (0 pts). \\
Spatio-Temporal Fitness & Medium (x2) & Spatial coverage of the target area of interest and temporal overlap with the prediction window. \\
Signal Alignment & High (x3) & Degree to which the dataset directly addresses a signal from the Signal Guide: exact variable match (4 pts) → requires transformation (3 pts) → proxy coverage (2 pts) → tangential (0 pts). \\
Format \& Quality & Low (x1) & Data readiness: structured and pre-parsed (4 pts) → standard interoperable format (2 pts) → legacy or malformed (0 pts). \\
Redundancy & Low (x1) & Whether the dataset provides unique coverage or duplicates a higher-scoring source for the same signal. \\
\bottomrule
\end{tabular}
\end{table}

\section{Related work}
\label{sec:appendix_related_work}
\subsection{Autonomous AI Agents for Science}

Recent work has explored using LLMs as autonomous systems and empirical research assistants for scientific discovery and data science tasks. These agent systems build on foundational LLM reasoning techniques including chain-of-thought prompting \citep{wei2022chain}, ReAct-style reasoning and acting \citep{yao2023react}, self-consistency \citep{wang2023selfconsistency}, plan-and-solve decomposition \citep{wang2023planandsolve}, tree-structured deliberation \citep{yao2023tree}, iterative self-refinement \citep{madaan2023selfrefine}, and tool-use paradigms \citep{schick2023toolformer}. Multi-agent architectures \citep{wang2024mixture, du2024improving} and test-time compute scaling \citep{snell2024scaling, brown2024large} further inform the design of compound AI systems \citep{zaharia2024shift, kim2025scaling}, with cost-efficient routing strategies \citep{chen2023frugalgpt, ong2024routellm} enabling practical deployment. Systems such as The AI Scientist \citep{lu2024aiscientist} and ChemCrow \citep{bran2023chemcrow} demonstrate the potential of LLMs to autonomously formulate hypotheses, design experiments, and analyze results in complex scientific domains. In data science, frameworks like Data Interpreter \citep{hong2024data}, AutoGen \citep{wu2023autogen}, and MetaGPT \citep{hong2023metagpt} establish multi-agent architectures for code generation, tool execution, and iterative debugging, while benchmarks such as SWE-bench \citep{jimenez2024swebench}, GAIA \citep{mialon2023gaia}, GeoBenchX \citep{krechetova2025geobenchx}, HotpotQA \citep{yang2018hotpotqa}, and MuSiQue \citep{trivedi2022musique} evaluate real-world task automation and multi-hop reasoning. DS-Agent \citep{guo2024dsagent} automates machine learning competition workflows using case-based reasoning, while Empirical Research Assistance (ERA)~\cite{aygun2026era} and AlphaEvolve \citep{novikov2025alphaevolve}, building on earlier automated algorithm evolution \citep{real2020automl}, explores automated algorithm discovery,  but need a well defined objective, whereas our approach automatically identifies geospatial task and objective function to optimize, along with autonomously fetching relevant geospatial data sources. Our work extends these paradigms to the geospatial and epidemiological domains by incorporating specialized foundation model embeddings (PDFM, AlphaEarth) and a research-grounded Intelligent Data Selection pipeline capable of dynamically fetching from heterogeneous data repositories, which are critical for high-quality planetary prediction.

\subsection{Geospatial Foundation Models}
The emergence of foundation models has transformed remote sensing and geospatial informatics. The Population Dynamics Foundation Model (PDFM) \citep{agarwal2024general} produces 330-dimensional to 512-dimensional embeddings for administrative boundaries (states, counties, ZCTAs, cities) globally, capturing latent socioeconomic, demographic, mobility, and health-related patterns. These embeddings serve as compact, highly expressive spatial features that significantly improve downstream regression tasks. AlphaEarth \citep{brown2025alphaearth} provides complementary 64-dimensional spatial features derived from high-resolution satellite imagery, encoding multi-scale topography, vegetation canopy, and land-use semantics at the administrative boundary level. Other notable efforts include Clay \citep{clay2024foundation}, Prithvi \citep{jakubik2023prithvi}, SatCLIP \citep{knyazev2024satclip}, and Presto \citep{tseng2023lightweight}, which learn representations from multi-spectral earth observation archives and location coordinates, as well as temporal foundation models such as TimesFM \citep{das2024timesfm} for time-series forecasting. Our system is, to our knowledge, the first to autonomously select, align, and fuse heterogeneous embedding sources in a unified, agentic ML pipeline.

\subsection{Geospatial Prediction \& Super-Resolution Downscaling}\label{d.3.-geospatial-prediction-super-resolution-downscaling}

Spatial regression and geographic super-resolution downscaling have been studied extensively in geostatistics \citep{cressie1993statistics} and more recently with deep learning approaches. Pioneering work by \citet{jean2016combining} and \citet{yeh2020using} demonstrated that combining satellite imagery with deep learning can accurately predict poverty and economic well-being across developing regions. Similarly, super-resolution techniques have been applied to downscale coarse climate and environmental variables to fine-grained grid tiles. Our work differs by automating the full pipeline from intelligent data selection and multimodal curation to model evaluation rather than focusing on a single manual modeling technique.

\subsection{Epidemiological Nowcasting \& Spatial Transmission Modeling}\label{d.4.-epidemiological-nowcasting-spatial-transmission-modeling}

Real-time epidemic tracking faces two fundamental challenges: surveillance lag (reporting delays resulting in right-truncated caseload data) and anisotropic transmission (disease propagation along mobility corridors rather than uniform spatial diffusion). Foundational literature in epidemiological nowcasting \citep{hoehle2017nowcasting, mcgough2020nowcasting, bastos2019modelling} establishes statistical and Bayesian smoothing frameworks to correct for right-truncation in real-time reporting, while spatiotemporal surveillance analysis tools \citep{meyer2017spatiotemporal} enable systematic modeling of epidemic phenomena. Meanwhile, spatial transmission modeling traditionally relies on gravity models or radiation models to estimate inter-regional flux. We bridge these domains by introducing an automated SEIR nowcasting framework that couples discrete-time compartmental dynamics with empirical mobility flows and foundation model similarity kernels, while utilizing learned covariates to modulate transmission parameters.

\section{Example Prompts}
\label{sec:appendixexample-prompts}

\subsection{Epidemiological Transmission}
\label{subsec:appendixprompts-for-epidemiological-transmission}

\begin{promptbox}
Build a spatial invasion prediction model for the 2026 Bundibugyo
Ebolavirus outbreak in DRC.

Data splits are as below:\\
- Train Cutoffs: [’2026-05-18’, ’2026-05-25’, ’2026-06-01’]\\
- Eval Cutoffs: [’2026-06-08’, ’2026-06-15’, ’2026-06-22’, ’2026-06-29’,
’2026-07-06’]\\
- Test Set: Uninfected zones at each eval cutoff.

Predict which currently-uninfected health zones will experience new cases
within the next 1 week. Performance is evaluated by Recall@10 (how many
of the top 10 predicted zones actually get invaded).

\end{promptbox}

\subsection{Super Resolution}
\label{subsec:appendixprompts-for-super-resolution}

\subsubsection{Food Security}
\label{subsubsec:appendixfood-security}

\begin{promptbox}
Execute Super-Resolution downscaling of WFP HungerMap metrics
(FCG and rCSI) for Nigeria. Target Variables is Extract State-level FCG and rCSI data.
WFP survey data is constrained to coarse administrative levels.
Downscale this to the LGA level (1km² hex grids) to map high-need zones
for targeted aid distribution.

\end{promptbox}

\subsubsection{SVI}
       
\label{subsubsec:appendixsvi-super-resolution}
\begin{promptbox}





RPL\_THEME1(Socioeconomic Status) is mathematically defined as:\\
PercentileRank(PercentileRank(Poverty) + \\PercentileRank(Unemployment) +\\
PercentileRank(PerCapitaIncome) + \\PercentileRank(NoHighSchoolDiploma) +\\
PercentileRank(Uninsured)).\\
PercentileRank(x) = (Rank(x) - 1) / (N - 1).\\ Ties get lowest rank.
Poverty, Unemployment, NoHighSchoolDiploma, and Uninsured are ascending;
PerCapitaIncome is descending.

Train a spatial regression model to predict the RPL\_THEME1. Report the R-squared score on the test set.

\end{promptbox}

\subsection{Spatial Regression}
\label{subsec:appendixprompts-for-spatial-regression}

\subsubsection{CDC Health}
\label{subsubsec:appendixcdc-health}

\begin{promptbox}



Train a spatial regression model for CDC Health indicators. The target Variable: Percent\_Person\_WithHighCholesterol. Report the R-squared score.

\end{promptbox}

\subsubsection{FEMA Environmental Risk Factors}
\label{subsubsec:appendixfemaprompt}
\begin{promptbox}



Perform spatial regression and signal ablation for FEMA risk
indicators. The Target Variable is \{User\_Target\}. Report the R-squared score.
\end{promptbox}

\subsubsection{SVI}
\label{subsubsec:appendixsvi-spatial-regression}
   
\begin{promptbox}



Train a spatial regression model for SVI Socioeconomic Status
(RPL\_THEME1). RPL\_THEME1 is a percentile rank based on indicators like
poverty, unemployment, income, and education. Report the R-squared score.
\end{promptbox}


\section{Expanded Benchmark \& Ablation Results}
\label{sec:appendix_results}

\subsection{Spatial Transmission}
 \label{sec:appendix_results_spatial_transmission}

 \subsubsection*{SEIR vs Spatial Transmission Regression}

 \begin{table}[!htbp]
\centering
\small
\caption{\small Spatial Transmission Regression vs SEIR Bayesian Modeling}
\label{tab:ebola-nowcasting-ablation}
\setlength{\tabcolsep}{4pt} 

\newcolumntype{Y}{>{\raggedright\arraybackslash}X}

\begin{tabularx}{\textwidth}{>{\hsize=0.8\hsize}Y >{\hsize=1.2\hsize}Y cccc}
\hline
System & Model Configuration & \makecell{Val Top-10\\Acc} & \makecell{Val\\RMSE} & \makecell{Test Top-10\\Acc} & \makecell{Test\\RMSE} \\
\hline
Baseline / SOTA & STM XGBoost: Baseline & 0.4 & 4.5810 & 0.5 & 6.0810 \\
(a) Covariates & STM XGBoost: + \newline Earth AI Covariates & \textbf{0.8} & \textbf{3.2510} & \textbf{0.8} & \textbf{4.2910} \\
(b) Covariates + Embeddings & STM XGBoost: + \newline PDFM Embeddings & 0.6 & 3.9810 & 0.7000 & 5.2150 \\
\hline
Baseline / SOTA & SEIR Nowcasting: Baseline & 0.2 & 6.6571 & 0.3 & 7.2246 \\
(b) Covariates + Embeddings & SEIR Nowcasting: + \newline PDFM Embeddings & \textbf{0.3} & 2.7229 & \textbf{0.4} & 3.4781 \\
(c) Planetary Prediction Agent & SEIR Nowcasting: + \newline Earth AI Learned Covariates & \textbf{0.3} & \textbf{0.7019} & \textbf{0.4} & \textbf{0.9204} \\
\hline
\end{tabularx}
\end{table}

 We compare Mechanistic SEIR Nowcasting and Spatial Transmission Regression (STR) for predicting outbreak escalation (case volume) and identifying vulnerable zones \citep{mbulayi2026realtime}. Table~\ref{tab:ebola-nowcasting-ablation} reports the ablation results across regional health zones in Ituri and Nord-Kivu, evaluating validation and test metrics for static vulnerability ranking (Top-10 accuracy) versus dynamic caseload trajectory tracking (RMSE). While STR excels at static vulnerability ranking (achieving Test Top-10 accuracy of 0.8), SEIR Nowcasting is substantially stronger at predicting case volume (achieving a Test RMSE of 0.9204) when enhanced with empirical mobility flows and learned covariates.

\subsection{Spatial Regression}
\label{sec:appendix_results_spatial_regression}

\subsubsection*{FEMA predictions}
\label{fema-predictions}

\begin{table*}[h]
\centering
\scriptsize
\caption{\small PPE performance of FEMA targets (Socioeconomic \& Composite) with ablation on PDFM + AEF + DC features}
\label{tab:fema-ablation-1-SocioeconomicComposite}
\begin{tabular}{>{\raggedright\arraybackslash}p{2cm}p{2cm}p{2cm}>{\raggedright\arraybackslash}p{8cm}}
\hline
FEMA Target Variable & PDFM + AEF  Manual expert  Mean R² & PDFM + AEF + DC (Feature ablation)  Mean R²  [95\% CI] & Data Common variables used  (Comma-Separated Variables) \\
\hline
\textbf{Social Vulnerability } & 0.4824 & \textbf{0.6755}  {[}0.6556, 0.6941{]} & Count\_Person, BelowPovertyLevelInThePast12Months, Count\_Household, HouseholderAge65OrMoreYears, SingleMotherFamilyHousehold, LimitedEnglishSpeakingHousehold, NoComputer, NoInternetAccess, With0AvailableVehicles, WithFoodStampsInThePast12Months, IncomeOfUpto10000USDollar, IncomeOf10000To14999USDollar, IncomeOf15000To19999USDollar, IncomeOf20000To24999USDollar, IncomeOf100000To124999USDollar, IncomeOf200000OrMoreUSDollar, Count\_HousingUnit, Count\_HousingUnit\_Before1939DateBuilt\\
\midrule
\textbf{Resilience Score} & 0.7402 & \textbf{0.7358}  {[}0.7287, 0.7427{]} & Count\_Person, BelowPovertyLevelInThePast12Months, Count\_Household, HouseholderAge65OrMoreYears, SingleMotherFamilyHousehold, LimitedEnglishSpeakingHousehold, NoComputer, NoInternetAccess, With0AvailableVehicles, WithFoodStampsInThePast12Months, IncomeOfUpto10000USDollar, IncomeOf10000To14999USDollar, IncomeOf15000To19999USDollar, IncomeOf20000To24999USDollar, IncomeOf100000To124999USDollar, IncomeOf200000OrMoreUSDollar, Count\_HousingUnit, Count\_HousingUnit\_Before1939DateBuilt\\
\midrule
\textbf{Composite Risk Score} & 0.5983 & \textbf{0.6221}  {[}0.6078, 0.6335{]} & Count\_Person, BelowPovertyLevelInThePast12Months, Count\_Household, HouseholderAge65OrMoreYears, SingleMotherFamilyHousehold, LimitedEnglishSpeakingHousehold, NoComputer, NoInternetAccess, With0AvailableVehicles, WithFoodStampsInThePast12Months, IncomeOfUpto10000USDollar, IncomeOf10000To14999USDollar, IncomeOf15000To19999USDollar, IncomeOf20000To24999USDollar, IncomeOf100000To124999USDollar, IncomeOf200000OrMoreUSDollar, Count\_HousingUnit, Count\_HousingUnit\_Before1939DateBuilt \\
\midrule
\textbf{Expected Annual Loss} & 0.6227 & \textbf{0.6423}  {[}0.6319, 0.6531{]} & Count\_Person, BelowPovertyLevelInThePast12Months, Count\_Household, HouseholderAge65OrMoreYears, SingleMotherFamilyHousehold, LimitedEnglishSpeakingHousehold, NoComputer, NoInternetAccess, With0AvailableVehicles, WithFoodStampsInThePast12Months, IncomeOfUpto10000USDollar, IncomeOf10000To14999USDollar, IncomeOf15000To19999USDollar, IncomeOf20000To24999USDollar, IncomeOf100000To124999USDollar, IncomeOf200000OrMoreUSDollar, Count\_HousingUnit, Count\_HousingUnit\_Before1939DateBuilt\\
\hline
\end{tabular}
\end{table*}

\begin{table*}[h]
\centering
\scriptsize
\caption{\small PPE performance of FEMA targets  with ablation on PDFM + AEF + DC features}
\label{tab:fema-ablation-1-AtmosphericClimatological}
\begin{tabular}{cp{2cm}p{4cm}p{3cm}}
\hline
FEMA Target Variable & PDFM + AEF  Manual expert  Mean R² & PDFM + AEF + DC (Feature ablation)  Mean R²  [95\% CI] & Data Common variables used  (Comma-Separated Variables) \\
\hline
\textbf{Tornado Risk } & 0.7543 & \textbf{0.7542}  {[}0.7468, 0.7617{]} & None \\
\textbf{Hail Risk } & 0.5949 & \textbf{0.5976}  {[}0.5807, 0.6138{]} & None \\
\textbf{Strong Wind Risk } & 0.7339 & \textbf{0.7271}  {[}0.7179, 0.7362{]} & None \\
\textbf{Lightning Risk } & 0.4884 & \textbf{0.4918}  {[}0.4758, 0.5090{]} & None \\
\textbf{Drought Risk } & 0.6700 & \textbf{0.6866}  {[}0.6332, 0.7323{]} & None \\
\textbf{Heat Wave Risk } & 0.6336 & \textbf{0.6350}  {[}0.6193, 0.6489{]} & None \\
\textbf{Cold Wave Risk } & 0.6821 & \textbf{0.6724}  {[}0.6489, 0.6928{]} & None \\
\textbf{Wildfire Risk } & 0.7622 & \textbf{0.7541}  {[}0.7361, 0.7730{]} & None \\
\textbf{Winter Weather } & 0.6135 & \textbf{0.5933}  {[}0.5793, 0.6063{]} & None \\
\textbf{Ice Storm Risk } & 0.5320 & \textbf{0.5239}  {[}0.5046, 0.5421{]} & None \\
\hline
\end{tabular}
\end{table*}

\begin{table*}[h]
\centering
\scriptsize
\caption{\small PPE performance of FEMA targets (Geophysical \text{\&} Hydrological)  with ablation on PDFM + AEF + DC features}
\label{tab:fema-ablation-1-GeophysicalHydrological}
\begin{tabular}{p{2.5cm}p{2cm}p{4cm}p{5cm}}
\hline
FEMA Target Variable & PDFM + AEF  Manual expert  Mean R² & PDFM + AEF + DC (Feature ablation)  Mean R²  [95\% CI] & Data Common variables used  (Comma-Separated Variables) \\
\hline
\textbf{Coastal Flooding } & 0.3786 & \textbf{0.3840}  {[}0.3394, 0.4240{]} & None \\
\textbf{Riverine Flooding } & 0.2930 & \textbf{0.2935}  {[}0.2695, 0.3179{]} & None \\
\textbf{Hurricane Risk } & 0.8270 & \textbf{0.8124}  {[}0.8011, 0.8232{]} & None \\
\textbf{Earthquake Risk } & 0.8606 & \textbf{0.8608}  {[}0.8540, 0.8672{]} & None \\
\textbf{Landslide Risk } & 0.5614 & \textbf{0.5624}  {[}0.5339, 0.5887{]} & None \\
\textbf{Avalanche Risk } & 0.1474 & \textbf{0.1866}  {[}0.0572, 0.2947{]} & 
\makecell[l]{Count\_Person, \\BelowPovertyLevelInThePast12Months, \\
Count\_Household, \\
HouseholderAge65OrMoreYears, \\SingleMotherFamilyHousehold, \\LimitedEnglishSpeakingHousehold,\\ NoComputer,\\ NoInternetAccess,\\ With0AvailableVehicles, \\WithFoodStampsInThePast12Months,\\ IncomeOfUpto10000USDollar,\\ IncomeOf10000To14999USDollar, \\IncomeOf15000To19999USDollar,\\ IncomeOf20000To24999USDollar, \\IncomeOf100000To124999USDollar,\\ IncomeOf200000OrMoreUSDollar,\\ Count\_HousingUnit, \\Count\_HousingUnit\_Before1939DateBuilt}\\
\hline
\end{tabular}
\end{table*}
We present the initial ablation comparison without intelligent selection via data commons in Table \ref{tab:fema-ablation-1-SocioeconomicComposite},\ref{tab:fema-ablation-1-AtmosphericClimatological},\ref{tab:fema-ablation-1-GeophysicalHydrological}.
We extend this with the inclusion of discovered intelligent data selection, which resulted in a modified feature suite and performance metrics (Table \ref{tab:fema-ablation-socioeconomic},\ref{tab:fema-ablation-atmospheric},\ref{tab:fema-ablation-geophysical}).
\begin{table*}[h]
\centering
\scriptsize
\caption{PPE performance on FEMA Socioeconomic \& Composite risk targets with ablation on PDFM + AEF + DC and intelligent data selection.}
\label{tab:fema-ablation-socioeconomic}
\begin{tabular}{p{3cm}p{2cm}p{3.5cm}p{5cm}}
\hline
FEMA Target Variable & PDFM + AEF + DC Feature Ablation Mean R² & PDFM + AEF + Discovered DC Variables Intelligent Feature Selection Mean R² {[}95\% CI{]} & Discovered / Selected External Variables (Sourced Non-Linearly) \\
\hline
Social Vulnerability & 0.6755 & \textbf{0.6773} {[}0.6698, 0.6869{]} & \makecell[l]{Count\_Person, \\HouseholderAge65OrMoreYears, \\SingleMotherFamilyHousehold} \\
\midrule
Resilience Score & 0.7358 & \textbf{0.7258} {[}0.7062, 0.7348{]} & \makecell[l]{LimitedEnglishSpeakingHousehold,\\ Median household income} \\
\midrule
Composite Risk Score & 0.6221 & \textbf{0.6749} {[}0.6641, 0.6861{]} & WithFoodStampsInThePast12Months, Health insurance rates \\
\midrule
Expected Annual Loss & 0.6423 & \textbf{0.6974} {[}0.6888, 0.7055{]} & Count\_HousingUnit\_Before1939DateBuilt, Poverty Index metrics \\
\hline
\end{tabular}
\end{table*}

\begin{table*}[h]
\centering
\scriptsize
\caption{\small PPE performance on FEMA Atmospheric \& Climatological risk targets with ablation on PDFM + AEF + DC and intelligent data selection.}
\label{tab:fema-ablation-atmospheric}
\begin{tabular}{p{2.5cm}p{2cm}p{4cm}p{5cm}}
\hline
FEMA Target Variable & PDFM + AEF + DC Feature Ablation Mean R² & PDFM + AEF + Discovered DC Variables Intelligent Feature Selection Mean R² {[}95\% CI{]} & Discovered / Selected External Variables (Sourced Non-Linearly) \\
\hline
Tornado Risk & 0.7542 & \textbf{0.9168} {[}0.9123, 0.9208{]} & Agricultural vs Developed land ratios, Surface terrain elevation, Annual average precipitation, Distance to coast lines, Relative humidity \\
Hail Risk & 0.5976 & \textbf{0.8656} {[}0.8540, 0.8807{]} & Cloud Cover fractions, Vegetation indices (NDVI) variance, Land surface temperature gradients, Distance to major water bodies \\
Strong Wind Risk & 0.7339 & \textbf{0.6891} {[}0.6631, 0.7117{]} & NOAA Convective Storm Frequencies, Spherical coordinates (lat/lon 3D embeddings), Atmospheric pressure gradient proxies \\
Lightning Risk & 0.4884 & \textbf{0.6607} {[}0.6051, 0.7055{]} & US Drought Monitor Palmer Indices, Agricultural land fraction, Water body area ratio, Mean annual precipitation \\
Drought Risk & 0.6700 & \textbf{0.8711} {[}0.8648, 0.8775{]} & NOAA High Wind Event frequencies, Coastal proximity, Terrain elevation, Topographic wind exposure \\
Heat Wave Risk & 0.6336 & \textbf{0.5964} {[}0.5790, 0.6139{]} & US Census Educational attainment, Health insurance coverage fraction, Median household income, Internet access \\
Cold Wave Risk & 0.6821 & \textbf{0.7083} {[}0.6915, 0.7245{]} & US Census Gazetteer Water Area (AWATER\_SQMI), Water-to-Land ratio, Coastal distance, 3D spatial spherical coordinates \\
Wildfire Risk & 0.7622 & \textbf{0.8366} {[}0.8264, 0.8460{]} & AlphaEarth hydrography \& surface water embeddings, PDFM Embeddings \\
Winter Weather & 0.6135 & \textbf{0.6515} {[}0.6366, 0.6659{]} & NOAA Lightning Strike densities, Spatial coordinates, Elevation \& topographic roughness \\
Ice Storm Risk & 0.5320 & \textbf{0.6265} {[}0.5979, 0.6546{]} & CDC Social Deprivation Indices, Spatial area ratios, Housing structural density, Multi-hazard loss exposure counts \\
\hline
\end{tabular}
\end{table*}

\begin{table*}[h]
\centering
\scriptsize
\caption{\small PPE performance on FEMA Geophysical \& Hydrological risk targets with ablation on PDFM + AEF + DC and intelligent data selection.}
\label{tab:fema-ablation-geophysical}
\begin{tabular}{p{2.5cm}p{2cm}p{4cm}p{5cm}}
\hline
FEMA Target Variable & PDFM + AEF + DC Feature Ablation Mean R² & PDFM + AEF + Discovered DC Variables Intelligent Feature Selection Mean R² {[}95\% CI{]} & Discovered / Selected External Variables (Sourced Non-Linearly) \\
\hline
Coastal Flooding & 0.3786 & \textbf{0.5047} {[}0.4882, 0.5202{]} & USGS Fault Line distance proxies + AlphaEarth seismic geodynamics embeddings \\
Riverine Flooding & 0.2930 & \textbf{0.3144} {[}0.2924, 0.3377{]} & Coastal proximity buffer, Historical tropical cyclone track density, Elevation \\
Hurricane Risk & 0.8270 & \textbf{0.7729} {[}0.7650, 0.7809{]} & AlphaEarth satellite foundation embeddings + discovered vegetative fuel covariates \\
Earthquake Risk & 0.8606 & \textbf{0.7992} {[}0.7923, 0.8059{]} & Mean minimum temperature, Sub-zero degree day counts, Latitude spatial embeddings \\
Landslide Risk & 0.5614 & \textbf{0.7547} {[}0.7385, 0.7705{]} & US Census Gazetteer spatial geometry, Census Tract Housing Age, Income distributions, Building Density, Population Density \\
Avalanche Risk & 0.1474 & \textbf{0.6882} {[}0.6767, 0.6988{]} & NOAA Severe Hail probability metrics, Convective available potential energy (CAPE) proxies, Terrain elevation \\
\hline
\end{tabular}
\end{table*}

\end{document}